\pdfoutput=1

\documentclass[11pt]{article}

\usepackage{emnlp2021}

\usepackage{gb4e}\noautomath
\usepackage{times}
\usepackage{latexsym}
\usepackage{amsmath}
\usepackage{graphicx}
\usepackage{booktabs}
\usepackage{subfig}
\usepackage{tikz}
\usepackage{tikz-qtree}
\usepackage{tikz-qtree-compat}
\usepackage[linguistics]{forest}
\usepackage{cancel}
\usepackage{multirow}
\usepackage{pifont}
\usepackage[T1]{fontenc}

\usepackage[utf8]{inputenc}
\usepackage{CJKutf8}

\usepackage{microtype}

\usepackage{cleveref}
\usepackage{enumitem}
\newcommand{\suite}[1]{\textsc{#1}}
\newcommand{\eng}[1]{(#1)}
\newcommand{\modifier}[1]{#1:}

\title{Controlled Evaluation of Grammatical Knowledge in\\Mandarin Chinese Language Models}

\author{Yiwen Wang$^1$ \quad Jennifer Hu$^2$ \quad Roger Levy$^2$ \quad Peng Qian$^2$\\
  $^1$ John A. Paulson School of Engineering
and Applied Sciences, Harvard University\\
  $^2$ Department of Brain and Cognitive Sciences, MIT \\
  \texttt{ yiwenwang@g.harvard.edu \quad  \{jennhu,rplevy,pqian\}@mit.edu }
}

\begin{document}
\maketitle
\begin{abstract}
Prior work has shown that structural supervision helps English language models learn generalizations about syntactic phenomena such as subject-verb agreement. However, it remains unclear if such an inductive bias would also improve language models' ability to learn grammatical dependencies in typologically different languages. Here we investigate this question in Mandarin Chinese, which has a logographic, largely syllable-based writing system; different word order; and sparser morphology than English. We train LSTMs, Recurrent Neural Network Grammars, Transformer language models, and Transformer-parameterized generative parsing models on two Mandarin Chinese datasets of different sizes. We evaluate the models' ability to learn different aspects of Mandarin grammar that
assess syntactic and semantic relationships. We find suggestive evidence that structural supervision helps with representing syntactic state across intervening content and improves performance in low-data settings, suggesting that the benefits of hierarchical inductive biases in acquiring dependency relationships may extend beyond English.
\end{abstract}

\section{Introduction}

A rich collection of targeted linguistic evaluations has shown that neural language models can surprisingly learn many aspects of grammar from unlabeled linguistic input \citep[e.g.,][]{linzen-etal-2016-assessing,gulordava-etal-2018-colorless,warstadt2020blimp,hu-etal-2020-systematic,xiang2021climp}.
There is also growing evidence that explicit modeling of syntax helps neural network-based language models represent syntactic state and exhibit human-like processing behaviors of non-local grammatical dependencies, including number agreement \cite{kuncoro-etal-2018-lstms}, negative polarity licensing, filler-gap dependencies \cite{wilcox-etal-2019-structural, hu-etal-2020-systematic}, and garden-path effects \citep{futrell-etal-2019-neural,hu-etal-2020-systematic}. However, this line of research has focused primarily on the syntax of English. It is unclear to what extent structural supervision may help neural language models generalize for languages with differing typologies. Expanding these analyses beyond English has the potential to inform scientific questions about inductive biases for language acquisition, as well as practical questions about model architectures that approach language-independence \citep{bender_achieving_2011}.

Here, we perform a controlled case study of grammatical knowledge in Mandarin Chinese language models.
The orthography and grammar of Chinese provide a useful testing ground given the differences from English and other Indo-European languages (see, e.g., \citealt{shopen_language_1985,li_chinese_2015}). Whereas today's Indo-European languages like English generally use phone-based orthography, Chinese uses a logographic system where each character generally responds to a syllable. Most Mandarin Chinese words are one or two syllables, influencing the distribution of tokens. Grammatically, Chinese has almost no inflectional morphology, and corpus studies suggest that the average dependency length of Mandarin Chinese sentences is larger than that of English sentences \citep{jiang2015effects}, with potential implications for language modeling. On the one hand, the need to track input across long dependencies may make structural supervision more beneficial for Mandarin Chinese language models; on the other hand, the prevalence of these dependencies may make it easier for them to learn to maintain non-local information without explicitly modeling syntax. 
Other fine-grained differences in typology also affect the types of syntactic tests that can be conducted. For example, 
since relative clauses precede the head noun in Chinese (unlike in English), we can manipulate the distance of a verb--object dependency by inserting relative clauses in between.
These characteristics motivate our choice of Mandarin Chinese as a language for evaluating structurally supervised neural language models. 

We design six classes of Mandarin test suites covering a range of syntactic and semantic relationships, some specific to Mandarin and some comparable to English. 
We train neural language models with differing inductive biases 
on two datasets of different sizes, and compare models' performance on our targeted evaluation materials. While most prior work investigating syntactically guided language models has used Recurrent Neural Network Grammar models \citep{dyer2016recurrent} -- potentially conflating structural supervision with a particular parameterization -- this work further explores structured Transformer language models \citep{qian-etal-2021-structural}. Our results are summarized as follows. 
We find that structural supervision yields greatest performance advantages in low-data settings, in line with prior work on English language models.  
Our results also suggest a potential benefit of structural supervision in deriving garden-path effects induced by local classifier--noun mismatch, and in maintaining syntactic expectations across intervening content within a dependency relation. These findings suggest that the benefits of hierarchical inductive biases in acquiring dependency relationships may not be specific to English.

\section{Targeted Linguistic Evaluation}

Linguistic minimal pairs have been used to construct syntactic test suites in English \citep[e.g.,][]{linzen-etal-2016-assessing, marvin2018targeted,mueller2020crosslinguistic, davis2020recurrent, warstadt2020blimp} and other languages such as Italian, Spanish, French, and Russian \citep{ravfogel2018lstm, gulordava-etal-2018-colorless,an-etal-2019-representation,mueller2020crosslinguistic, davis2020recurrent}. 
A minimal pair is formed by two sentences that differ in grammaticality or acceptability but are otherwise matched in structure and lexical content. \ref{itm:minimal-pair-a} and \ref{itm:minimal-pair-b} form an example English minimal pair differing in subject-verb agreement:
\begin{enumerate}[label=(\textbf{\Alph*})]
\vspace{-0.5em}
\item \label{itm:minimal-pair-a}{The man \emph{drinks} coffee everyday.}\vspace{-0.5em}
\item \label{itm:minimal-pair-b}{The man \emph{drink} coffee everyday.}
\vspace{-0.5em}
\end{enumerate}
The two sentences differ only at the main verb `drinks'/`drink', which must agree with its subject `The man' in number. Since only the third-person singular form `drinks' agrees with the subject, \ref{itm:minimal-pair-a} is grammatical, whereas \ref{itm:minimal-pair-b} is not.

The closest work to ours is the corpus of Chinese linguistic minimal pairs \citep[CLiMP;][]{xiang2021climp}, which provides a benchmark for testing syntactic generalization of Mandarin Chinese language models. While CLiMP focuses on building a comprehensive challenge set, the current work performs controlled experiments to investigate the effect of structural supervision on language models' ability to learn syntactic and semantic relationships. Moreover, the test items in CLiMP are (semi)-automatically generated, which may result in semantically anomalous sentences and introduce noise into the evaluation phase. In contrast, we manually construct the items in our test suites to sound as natural as possible.

\section{Methods}

\subsection{Evaluation Paradigm}

The general structure of our evaluation paradigm follows that of \citet{wilcox-etal-2019-structural} and \citet{hu-etal-2020-systematic}. 
We use surprisal as a linking function between the language model output and human expectations \citep{hale-2001-probabilistic}. Surprisal is defined as the inverse log probability of a word ($w_i$) conditioned on the preceding words in the same context ($w_1…w_{i-1}$):
 $$\text{surprisal}(w_i) = \text{log } \frac{1}{p(w_i|w_1,…w_{i-1})}$$
Our test suites take the form of a group of hand-written, controlled sentence sets. Each sentence set, or \emph{test item}, contains at least two minimally differing sentences, and each sentence contains a \emph{stimulus prefix} and a downstream \emph{target region}. The content of the target region remains fixed across the sentence variants within the test item, while the content of the stimulus varies in a minimal manner that modulates the sentence's grammaticality or acceptability. 
The target region, where we measure the surprisal output of a certain model, is underlined in all the example items described in \Cref{sec:test-suites}.

For each test item, we measure success by computing the difference in surprisals assigned by the model to the target region, conditioned on the ungrammatical vs.~grammatical stimulus prefixes. If the model successfully captures the dependency, it should be less surprised at the grammatical target region than the ungrammatical one, leading to a positive surprisal difference (ungrammatical $-$ grammatical). If this criterion is satisfied, then the model achieves a success score of 1, and 0 otherwise. These binary scores are averaged over test suite items and/or classes to obtain accuracy scores.

\subsection{Test Suites} \label{sec:test-suites}
We organize our materials into six classes of test suites, each of which assesses models' knowledge of a particular linguistic phenomenon.\footnote{All code and data, including test suites, can be found at \url{https://github.com/YiwenWang03/syntactic-generalization-mandarin} 
} Within each class, we develop individual test suites with different types of modifiers that intervene between the two ends of the dependency (including the no modifier case). For each test suite, we manually construct $\sim$30 test items, taking care to maintain semantic plausibility wherever possible. For details about individual test suites, see \Cref{sec:ind_test_suites}. While here we present example items of a specific type of modifier to introduce each test suite, full set of examples with other type of modifiers are included in \Cref{sec:modifiers}.

The test suite classes include both syntactic and semantic dependencies, some of which do not exist in languages that have been the focus of targeted evaluation, and thus have not yet been explored. \suite{Missing Object} evaluates syntactic knowledge of argument structure, \suite{Subordination} and \suite{Garden Path subject/object} assess representation of syntactic state, and \suite{classifier--noun Compatibility} and \suite{verb--noun Compatibility} evaluate a combination of syntactic and semantic factors
\citep{Sergio}. Looking cross-linguistically, one class assesses a phenomenon present in Mandarin but not English (\suite{classifier--noun Compatibility}); two classes assess an expectation-violation phenomenon that is present in both Mandarin and English but arises from different sources (\suite{Garden Path subject/object}); and three classes assess phenomena present in both languages (\suite{verb--noun Compatibility}, \suite{Missing Object}, and \suite{Subordination}). 
 
For each phenomenon targeted by a given test suite class, the 
two components of the syntactic/semantic dependency often occur adjacently in a sentence. However, if a language model robustly represents the dependency, then it should maintain its expectations even when intervening content is present between the upstream and downstream ends of the dependency. We assess the robustness of the models' grammatical knowledge on each test suite class by inserting three commonly-used types of modifiers to create non-local dependencies: adjectives, subject-extracted relative clauses (SRCs) and their variants, and object-extracted relative clauses (ORCs). The resulting set of test suites is described in greater detail in the following sections. 

\subsubsection{Classifier--Noun Compatibility}  

Classifiers are a special class of words in Chinese languages which are obligatorily used with numerals in a noun phrase. Each specific classifier in Mandarin 
is only compatible with a set of noun references that is largely semantically delimited. The general classifier \begin{CJK*}{UTF8}{gbsn}``个''\end{CJK*}\eng{\textsc{cl}$_\textsc{general}$}, in contrast, is compatible with most nouns. In the \suite{classifier--noun Compatibility} test suites, we evaluate whether a model expects nouns from a semantically-compatible class over those from a semantically-incompatible class, given a specific classifier.

\begin{CJK*}{UTF8}{gbsn}
\footnotesize

\begin{exe}
\exi{(1.a)} \gll 孩子 听到 了 一 首 熟悉 的 \underline{歌曲\hspace{1mm}。}\\
        child hear \textsc{pst} one \textsc{cl}$_\textsc{song}$ familiar \textsc{de} \underline{song .}\\
\glt ``The child heard a familiar song.''
\exi{$*$(1.b)} \gll 孩子 听到 了 一 张 熟悉 的 \underline{歌曲\hspace{1mm}。}\\
        child hear \textsc{pst} one \textsc{cl}$_\textsc{album}$ familiar \textsc{de} \underline{song .}\\
\glt `The child heard a familiar song.''
\exi{(1.c)} \gll 孩子 听到 了 一 张 熟悉 的 \underline{专辑\hspace{1mm}。}\\
        child hear \textsc{pst} one \textsc{cl}$_\textsc{album}$ familiar \textsc{de} \underline{album .}\\
\glt ``The child heard a familiar album.''
\exi{$*$(1.d)} 

\gll  孩子 听到 了 一 首 熟悉 的 \underline{专辑\hspace{1mm}。}\\
        child hear \textsc{pst} one \textsc{cl}$_\textsc{song}$ familiar \textsc{de} \underline{album .}\\
\glt ``The child heard a familiar album.''

\end{exe}
\end{CJK*}

Example (1) shows a test item from the suite with adjectival modifiers. We also consider ORCs and SRCs as modifiers in this test suite class (see \Cref{sec:cn-test-suites}). Here we consider the classifier \begin{CJK*}{UTF8}{gbsn}``首''\end{CJK*}\eng{\textsc{cl}$_\textsc{song}$}, which is compatible with the noun \begin{CJK*}{UTF8}{gbsn}``歌曲''\end{CJK*}\eng{song} but not the noun \begin{CJK*}{UTF8}{gbsn}``专辑''\end{CJK*}\eng{album}, and the classifier \begin{CJK*}{UTF8}{gbsn}``张''\end{CJK*}\eng{\textsc{cl}$_\textsc{album}$}, which is compatible with the noun \begin{CJK*}{UTF8}{gbsn}``专辑''\end{CJK*} but not the noun \begin{CJK*}{UTF8}{gbsn}``歌曲''\end{CJK*}. The four variants (1.a-d) show four possible combinations of the two classifiers and the two nouns. Here the target region is the sentence-final noun together with the period. We also check that the two nouns compared within each test item have similar frequency in the training data.
We measure surprisals at the sentence-final noun and the period. A human-like language model should assign lower surprisals to the target regions in (1.a) and (1.c), the items with an appropriate classifier--noun pair, and high surprisals to the target regions in (1.b) and (1.d), the items with mismatched classifier--noun pairs. In other words, we evaluate four pair-wise comparisons to see whether they meet the following criteria: (1.b) > (1.a), (1.d) > (1.c), (1.d) > (1.a), and (1.b) > (1.c). We report mean accuracy averaged across all four pair-wise comparisons as a model's accuracy on a given test suite.

\subsubsection{Garden-Path Effects}

Garden-path effects are a class of phenomena in human sentence processing, where the incremental parsing state of a sentence prefix needs to be reanalyzed as the comprehender processes a downstream disambiguator region \citep{bever1970cognitive}. We construct a set of test suites evaluating whether models exhibit 
garden-path effects induced by locally mismatched classifier--noun pairs situated within a globally coherent sentence, inspired by previous human behavioral studies \citep{wu2018effects}.

To illustrate the classifier-induced garden-path effect, consider examples (2.a) and (2.b): 

\begin{CJK*}{UTF8}{gbsn}
\footnotesize
\begin{exe}
\exi{(2.a)} \gll 他 离开 了 那 间 朋友 \underline{开} 的 工厂\hspace{1mm}。\\
        he leave \textsc{pst} that \textsc{cl}$_\textsc{building}$ friend \underline{start} \textsc{de} factory\hspace{1mm}. \\

\exi{(2.b)} \gll 他 离开 了 那 个 朋友 \underline{开} 的 工厂\hspace{1mm}。\\
        he leave \textsc{pst} that \textsc{cl}$_\textsc{general}$ friend \underline{start} \textsc{de} factory\hspace{1mm}.\\
\glt ``He left the factory that the friend started.''
\end{exe}
\end{CJK*}

\noindent
In (2.b), the general classifier \begin{CJK*}{UTF8}{gbsn}``个''\end{CJK*}\eng{\textsc{cl}$_\textsc{general}$} is compatible with the immediately following noun, \begin{CJK*}{UTF8}{gbsn}``朋友''\end{CJK*}\eng{friend}, resulting in a garden-path interpretation that this noun is the object of the main-clause verb \begin{CJK*}{UTF8}{gbsn}``离开''\end{CJK*}\eng{leave}. This interpretation is disconfirmed, however, by the next verb, \begin{CJK*}{UTF8}{gbsn}``开''\end{CJK*}\eng{start}, which indicates that the noun \begin{CJK*}{UTF8}{gbsn}``朋友''\end{CJK*} is actually the start of a relative clause preceding the main-clause object, \begin{CJK*}{UTF8}{gbsn}``工厂''\end{CJK*}\eng{factory}. This means that the relative clause verb \begin{CJK*}{UTF8}{gbsn}``开''\end{CJK*} should be highly surprising in (2.b). In (2.a), in contrast, the specific classifier \begin{CJK*}{UTF8}{gbsn}``间''\end{CJK*}\eng{\textsc{cl}$_\textsc{building}$} is incompatible with \begin{CJK*}{UTF8}{gbsn}``朋友''\end{CJK*} (though it is compatible with \begin{CJK*}{UTF8}{gbsn}``工厂''\end{CJK*}),
cueing the upcoming relative clause structure \citep{wu2018effects}. A human-like language model should thus show a higher surprisal at the target verb \begin{CJK*}{UTF8}{gbsn}``开''\end{CJK*} for (2.b) than (2.a).

We design test suites similar to the structure of example (2) and (3). Examples (2.a-b) show the structure of items in the \suite{Garden Path object} set, where an ORC modifies the \emph{object} of the main clause verb and the target region is the verb immediately following the closest noun to the classifier. Examples (3.a-b) show the basic structure of items in the \suite{Garden Path subject} set, where an ORC modifies the sentence \emph{subject}. 

\begin{CJK*}{UTF8}{gbsn}
\footnotesize
\begin{exe}
\exi{(3.a)} \gll 那 间 朋友 开 \underline{的} 工厂 倒闭 了 。\\
        that \textsc{cl}$_{\textsc{building}}$ friend start \underline{\textsc{de}} factory close \textsc{pst} .\\

\exi{(3.b)} \gll 那 个 朋友 开 \underline{的} 工厂 倒闭 了 。\\
        that \textsc{cl}$_{\textsc{general}}$ friend start \underline{\textsc{de}} factory closed \textsc{pst} .\\
\glt ``The factory that the friend started has closed.''
\end{exe}
\end{CJK*}

\noindent 
Here in (3.b), the garden-path interpretation is that the ORC's subject \begin{CJK*}{UTF8}{gbsn}``朋友''\end{CJK*} may be initially analyzed as the subject of the sentence during incremental processing; the target region of the garden-path effect is the disambiguating word \begin{CJK*}{UTF8}{gbsn}``的''\eng{\textsc{de}}\end{CJK*} that ends the relative clause and precedes the head noun of the true main subject of the sentence, \begin{CJK*}{UTF8}{gbsn}``工厂''\end{CJK*}.

The criterion for getting a test item correct is that the model shows lower surprisal at the target region \begin{CJK*}{UTF8}{gbsn}``开'' \end{CJK*}\eng{start} in (2.a) than (2.b), and lower surprisal at the target region \begin{CJK*}{UTF8}{gbsn}``的''\end{CJK*} in (3.a) than (3.b). Examples (2) and (3) have no modifiers in between the classifier stimulus and the target region. To manipulate the length of dependency, we also consider same types of modifiers as in \suite{classifier--noun Compatibility} in the full test suites.

\subsubsection{Verb--Noun Compatibility}

Similar to \suite{classifier--noun Compatibility}, \suite{verb--noun Compatibility} is also a group of semantic test suites, assessing the consistency between a transitive verb\footnote{We test the transitivity of the verbs with a Tregex search \citep{levy-andrew-2006-tregex} on the CTB dataset, along with human judgments from native Mandarin speakers.} and its direct object noun. (4) shows an example with an adjective modifier in between the verb and its object, where (4.b) is semantically inconsistent since the word \begin{CJK*}{UTF8}{gbsn}``阅读''\end{CJK*}\eng{read} does not match the object noun
\begin{CJK*}{UTF8}{gbsn}``电脑''\end{CJK*}\eng{computer}. 

\begin{CJK*}{UTF8}{gbsn}
\footnotesize
\begin{exe}
\exi{(4.a)} \gll 我 修理 了 这 个 新 的 \underline{电脑 。} \\
        I fix \textsc{pst} this \textsc{cl} new \textsc{de} \underline{computer .}\\
        \glt ``I have fixed this new computer.''

\exi{$*$(4.b)} \gll 我 阅读 了 这 个 新 的 \underline{电脑 。}\\
        I read \textsc{pst} this \textsc{cl} new \textsc{de} \underline{computer .}\\
\glt ``I have read this new computer.''
\end{exe}
\end{CJK*}
The stimulus is the transitive verb, and the target region is the object noun and the period (which encapsulates the possibility of an incomplete but potentially grammatical sentence). We insert adjectives, ORCs and SRCs modifiers (same as in \suite{classifier--noun Compatibility}) between the verb and object. The expected behavior is that the surprisal at the target region being lower in the semantically consistent variant (here, (4.a)).

\subsubsection{Missing Object}
Next, we turn to phenomena primarily characterized by syntactic expectations. The first of these test suite classes, \suite{Missing Object}, assesses models' ability to track a direct object required by a transitive main verb.  
Consider (5.a) and (5.b) (no modifier case):

\begin{CJK*}{UTF8}{gbsn}
\footnotesize
\begin{exe}
\exi{(5.a)} \gll 记者 采访 了 科学家 \underline{。}\\
        journalist interview \textsc{pst} scientist \underline{.}\\
\glt ``The journalist interviewed the scientist.''

\exi{$*$(5.b)} \gll 记者 采访 了 \underline{。}\\
        journalist interview \textsc{pst} \underline{.}\\
\glt ``The journalist interviewed.''
\end{exe}
\end{CJK*}
(5.a) is grammatical and (5.b) is not, since the main verb \begin{CJK*}{UTF8}{gbsn}``采访''\end{CJK*}\eng{interview} requires a downstream direct object. To test if models learn this dependency, we record the model's surprisal at the sentence-final period \begin{CJK*}{UTF8}{gbsn}``。''\end{CJK*}. The model should be more surprised to see \begin{CJK*}{UTF8}{gbsn}``。''\end{CJK*} in (5.b) since it is less likely to end a sentence without an object needed by the verb. Note that this is a case where we assess human-likeness of an autoregressive language model by whether it is \emph{temporarily confused} as to the structural interpretation midway through the sentence, as evidenced by its next-word predictions.
 
To continue our investigation of long-distance dependencies, we add three types of modifiers: single SRC, coordinated SRCs, and embedded SRCs, exemplified in \Cref{sec:mobj-test-suites}, respectively.\footnote{We do not consider the conventional modifier types used in the other test suite classes because adding an ORC can be confounding --- the language model might be surprised at the ungrammatical target region due to the RC verb instead of the main transitive verb. Therefore, we instead focus on SRCs for this particular class.} We expect the insertion of modifiers after the main verb to increase difficulty, as the model must track the verb-object dependency over a greater amount of content. In addition, the parallel and hierarchical SRCs are longer and more syntactically complex than the simple SRCs.

\subsubsection{Subordination}

Finally, our \suite{Subordination} test suites assess the ability of a model to maintain global expectations for a main clause while inside a local subordinate clause. 
For example, consider (6.a) and (6.b) (no modifier case):

\begin{CJK*}{UTF8}{gbsn}
\footnotesize
\begin{exe}
\exi{(6.a)} \gll 如果 他 不 尝试 ， 他 将 失去 机会 \underline{。}\\
        if he \textsc{neg} try , he will lose opportunity \underline{.}\\
\glt ``If he doesn't try, he will lose the opportunity.''

\exi{$*$(6.b)} \gll 如果 他 不 尝试 \underline{。}\\
        if he \textsc{neg} try \underline{.} \\
\glt ``If he doesn't try.''
\end{exe}
\end{CJK*}
In this case, we test the surprisal at the sentence-final period. If the model correctly represents the gross syntactic state within the subordinate clause, then it should assign higher surprisal to the period in sentences like (6.b) than in sentences like (6.a). 
We include modifiers (same types as in \suite{classifier--noun Compatibility}) before the subject noun inside the matrix clause.

\subsection{Models}
To investigate the effect of syntax modeling in learning the dependencies described in \Cref{sec:test-suites}, we train four classes of neural language models by crossing two types of parameterization with two types of supervision. Two of our model classes are trained for vanilla next-word-prediction: Long Short-Term Memory networks \citep[LSTM;][]{HochSchm97} and Transformers \citep{attention}. The remaining two model classes are based on the LSTM and Transformer architectures, but explicitly incorporate syntactic structure during training: Recurrent Neural Network Grammars \citep[RNNG;][]{dyer2016recurrent} and Transformer-parameterized parsing-as-language-modelling models \citep[PLM;][]{qian-etal-2021-structural}. 
While prior work on structural supervision in English language models has focused primarily on RNNGs, both RNNGs and PLMs are joint probabilistic models of terminal word sequences along with the corresponding constituency parses. Thus, they both explicitly model syntax (in contrast to their vanilla language modeling counterparts), while featuring different parameterizations. 

We use the PyTorch implementation of the LSTM \cite{pytorch}. The Transformer and PLM models are based on the HuggingFace GPT-2 architecture \cite{transformer}. While we use the model architecture equivalent to the size of pre-trained GPT-2, we do not use the pre-trained tokenizer. All of our models are trained on a pre-tokenized Mandarin corpus and share the same vocabulary for each training dataset. Model sizes are reported in \Cref{tab:model} in \Cref{sec:model-info-ppl}.

For the LSTM and Tranformer models, we calculate the surprisal at the target region by taking the negative log of the model’s predicted conditional probability. We estimate the RNNGs and PLMs' word surprisals with word-synchronous beam search \citep{stern-etal-2017-effective}, following \citet{hale-etal-2018-finding} and \citet{wilcox-etal-2019-structural}. The action beam size is 100 and the word beam size is 10. For regions with multi-token content, we sum over the probabilities of each token.

As a baseline, we additionally implement an $n$-gram model with Kneser-Ney Smoothing \citep{kneser1995improved} using the SRILM toolkit \citep{stolcke2002srilm}. For cases where the smoothed $n$-gram model assigns identical probabilities to the target region across different conditions in a test item, we do tie-breaking by randomly flipping a fair coin to determine the outcome for that particular item.

\subsection{Corpus Data}
We consider two datasets to explore how training data size affects models ability to acquire grammatical knowledge \citep[similar to][]{hu-etal-2020-systematic}. The LSTM and Transformer models are trained on the raw text only, whereas the RNNG and PLM models are tained with additional syntactic annotations.

\paragraph{Chinese Treebank (CTB)}
The Chinese Treebank \citep[CTB 9.0;][]{ctb9} is a Chinese language corpus annotated with Penn Treebank-style \citep{Penn} constituency parses. 
We use the Newswire, Magazine articles, Broadcast news, Broadcast conversations, and Weblogs sections, as we expect these sources to contain well-formed sentences with a variety of syntactic constructions. 
We follow the split defined by \citet{shao2017character} to construct training, development, and test sets.

\paragraph{Xinhua News Data}
To investigate the effects of increased training data size on models' syntactic generalization, we create a larger corpus combining CTB with a subset from the Xinhua News corpus \citep{xinhua}.\footnote{
We include CTB to guarantee that this larger dataset covers the CTB vocabulary.} The Xinhua corpus contains metadata and content for 406K Mandarin news articles, collected from three mainstream media sources. Only the article contents are used for our training purposes.
The texts from Xinhua corpus are first split into sentences and then tokenized into words with SpaCy \citep{spacy}. We then obtain sentence parses with the Berkeley Neural parser \cite{kitaev-klein-2018-constituency, kitaev-etal-2019-multilingual}. 
We filter out extremely long sentences (>100 tokens) 
and map tokens occurring less than twice in the training data to fine-grained UNK tokens. 
\Cref{sec:corpus-stats} reports full statistics of the training corpora.

We train ten types of language models, crossing model architecture ($n$-gram, LSTM, RNNG, Transformer, and PLM) with dataset (Chinese Treebank and hybrid Xinhua dataset).\footnote{We denote the language models trained on the hybrid Xinhua dataset with the suffix ``-Xinhua''.} Each model type is trained with multiple random seeds,\footnote{We train RNNG models for 2 random seeds, and all other neural models for 3 random seeds. 
} and results are reported as averages across these instances. Model perplexity scores are reported in Table \ref{tab:ppl}.

\begin{figure*}
\includegraphics[width=\linewidth]{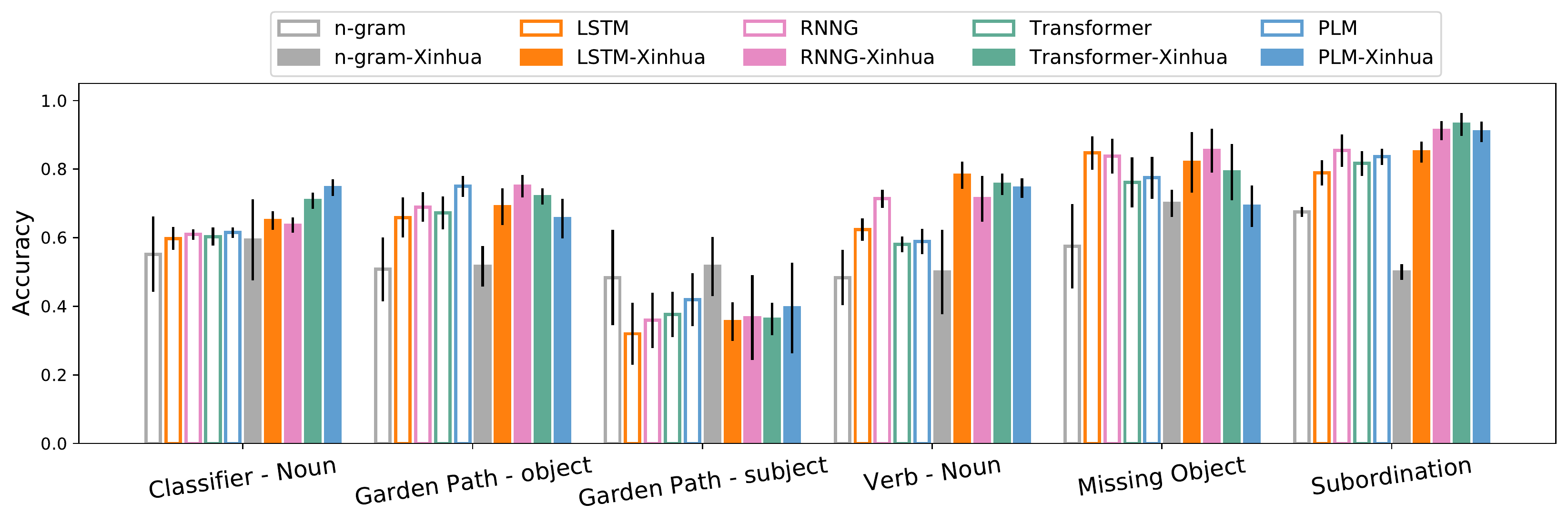}
    \caption{Accuracy by test suite class and model. For each test suite class, we average the accuracy across different modifier types (including the no modifier case). Error bars denote 95\% CIs of the mean accuracy score.}
    \label{fig:bymodelntest}
\end{figure*}

\begin{table*}[t]
\definecolor{g}{rgb}{0.0, 0.65, 0.31}
\definecolor{r}{rgb}{0.76, 0.23, 0.13}
\newcommand{\x}{{\normalsize\textcolor{r}{\ding{55}}}}
\renewcommand{\check}{{\normalsize\textcolor{g}{\ding{51}}}}
\centering
\scriptsize
\begin{tabular}{c|c|c|c|c}
\toprule
\multirow{2}{*}{} & \multicolumn{2}{c|}{CTB} & \multicolumn{2}{c}{Xinhua}\\

 & RNNG vs.~LSTM & PLM vs.~Transformer & RNNG vs.~LSTM & PLM vs.~Transformer \\ \midrule
 \suite{classifier--noun} & - & - & - & \check**\\
 \suite{Garden Path object} & - & \check**& - & \x*\\
 \suite{Garden Path subject} & - & - & - & - \\
 \suite{verb--noun} & \check**& - & \x* & - \\
 \suite{Missing Object} & - & - & - & \x*** \\
 \suite{Subordination} & \check**& - & \check** & - \\
\bottomrule

\end{tabular}
\caption{Comparison between language models that perform explicit syntax modeling (RNNGs and PLMs) and their vanilla counterparts (LSTMs and Transformers). \check{} represents statistically significant improvement in structurally supervised language models, and \x{} represents the opposite direction. *: $p\leq .05$, **: $p\leq .01$, ***: $p\leq .001$. }
\label{tab:model_compare}
\end{table*}

\section{Results}

We begin by reporting the overall performance of the models on the test suite classes introduced in \Cref{sec:test-suites}. Figure \ref{fig:bymodelntest} shows accuracy scores averaged across test suites within each class.\footnote{For numerical accuracy scores, see \Cref{tab:model_results} in \Cref{sec:table_acc}. For results on individual test suites, see \Cref{fig:append} in \Cref{sec:results_ind_test_suites}.} 
First, we note that the $n$-gram baseline overall performs the worst among all language models, which matches our expectations since syntactic dependencies beyond the 5-token window are difficult for the model to capture.

\begin{figure}[t]
\centering
\includegraphics[width=\linewidth]{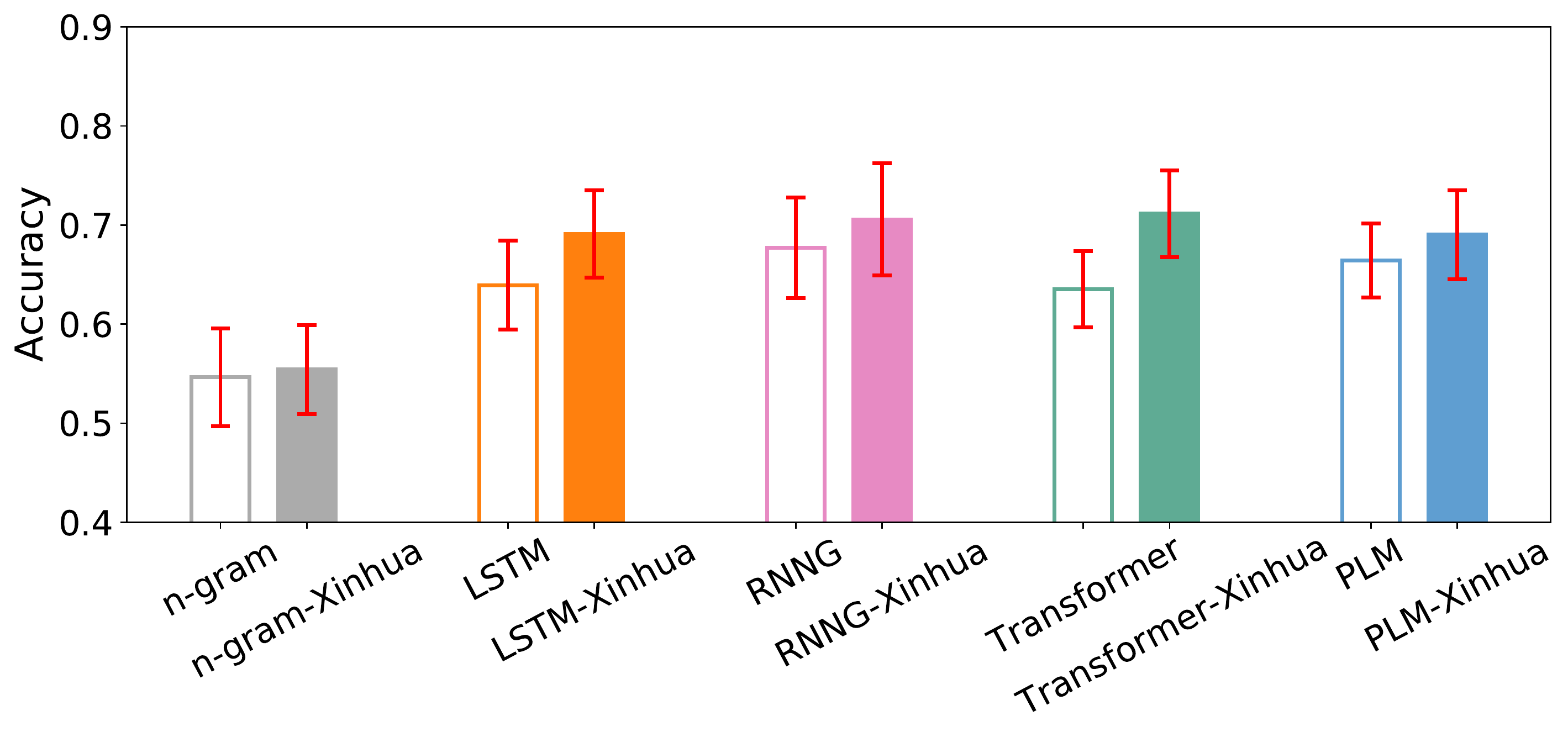}

\caption{Mean accuracy scores by model type across all test suites.}
\label{fig:acc_by_model_test}
\end{figure}

Turning to the neural models, we assess the effects of training data size and architecture by examining the mean accuracy scores across test suites for each model. We fit separate linear mixed-effects models comparing the effects of data size, using the \texttt{lme4} package in R \citep{lme4}. The dependent variable is the mean accuracy score (\Cref{fig:acc_by_model_test}). For each language model type, the main effect is a binary indicator of whether the model is trained on the CTB dataset or the larger Xinhua dataset. We include test suite class, modifier type, and model seed as random factors with random intercepts and slopes for the accuracy score. Across test suite classes, the Transformer-Xinhua models outperform their smaller CTB counterparts ($p<.05$), but the effect of data size is less clear for the LSTM, RNNG and PLM models. 

\begin{figure*}[t]
\centering
\includegraphics[width=\linewidth]{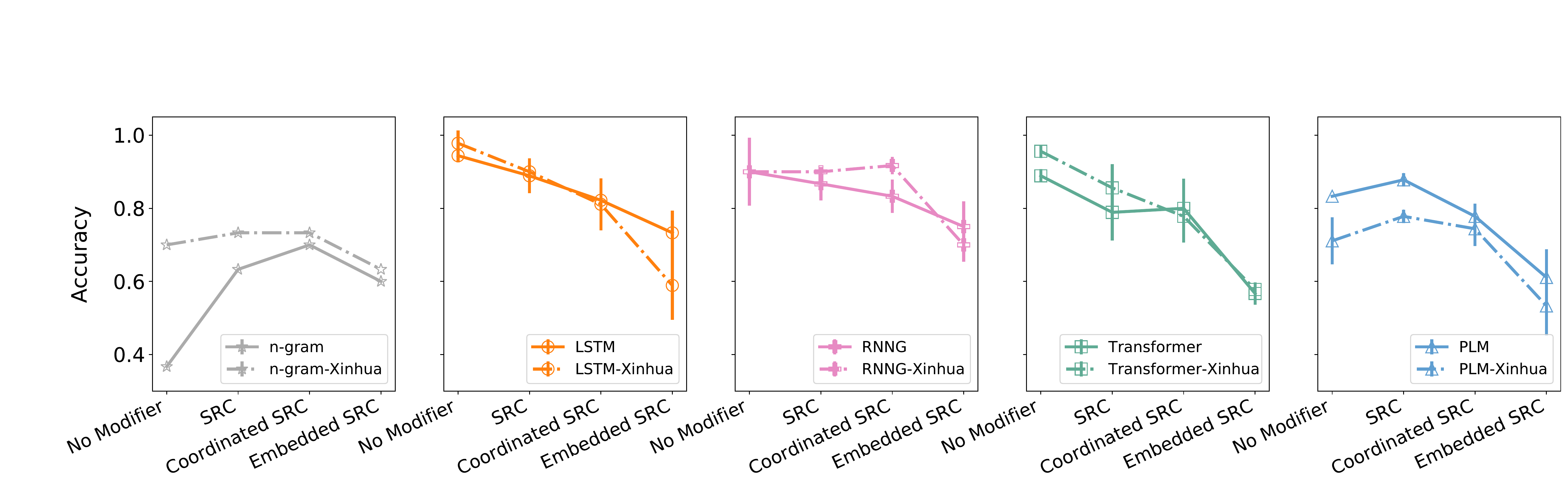}
\caption{Accuracy on \suite{Missing Object} as function of modifier complexity, for each model class.}
\label{fig:mobj_line}
\end{figure*}

Comparing different model architectures,
we find that in the \suite{Subordination} test suite class, RNNGs trained on the smaller CTB dataset achieve comparable performance to LSTMs trained on the larger Xinhua dataset ($p=.938$; linear mixed-effects model with model type as the main factor). 
Furthermore, for both \suite{verb--noun Compatibility} and \suite{Subordination}, the RNNGs perform better than the LSTMs and Transformers when trained on the smaller CTB dataset (see \Cref{fig:bymodelntest}).  These results suggest that an inductive bias for learning hierarchical structure may help in low-data Chinese settings, consistent with prior experiments conducted in English language models \citep{hu-etal-2020-systematic}.

Overall, we find only suggestive benefits of structural supervision. While prior work in English \citep{kuncoro-etal-2018-lstms,wilcox-etal-2019-structural,hu-etal-2020-systematic,qian-etal-2021-structural} has shown that RNNGs and PLMs can improve upon their vanilla LSTM and Transformer counterparts, 
the improvement is relatively smaller for Mandarin. To compare the performance of models with and without structural supervision, we fit a generalized linear mixed-effect model on the binary outcome (whether or not the model predicts a positive surprisal difference) for each test item, within each combination of test suite class, training data, and model parameterization. We consider a binary indicator of whether or not the model performs syntax modeling explicitly as the main factor, and include test item, modifier type and model seed as random factors. \Cref{tab:model_compare} summarizes the results. For the CTB-only models, the structurally supervised models (RNNG and PLM) achieve accuracy either significantly greater or comparable to the corresponding vanilla models (LSTM and Transformer) across all test suite classes. However, the pattern is less clear for the Xinhua-trained models: the structurally supervised models lead to both gains and losses in accuracy compared to the vanilla models. We conjecture that word segmentation and parsing errors in the automatically-annotated Xinhua dataset might have affected the learning process of the model. In addition, the Xinhua training data explored in this work is still not very large in size (<10 million tokens), so it could be that 
further benefits of syntactic supervision may be more pronounced with much larger training datasets. Nevertheless, the suggestive benefits of explicitly modelling syntax with very small amounts of data could have implications for language modeling in low-resource settings.

We also assess whether our models are better at capturing syntactic or semantic relationships. To do this, we group the six classes of test suites into three categories: syntactic dependency (\suite{Missing Object} and \suite{Subordination}), semantic relationship (\suite{classifier--noun Compatibility} and \suite{verb--noun Compatibility}), and a hybrid capturing semantically-driven syntactic state representations (\suite{Garden Path subject} and \suite{Garden Path object}). We find that the average accuracy score is higher in the syntactic test suites than in the semantic test suites ($p<.001$%
),\footnote{See \Cref{sec:compare-syntactic-semantic} for details.}
suggesting that the language models in our study -- including those with no explicit syntax modeling -- find it easier to learn syntactic dependencies than semantic relationships.

\subsection{Robustness to Intervening Content}

Next, we investigate the effect of structural supervision on tracking dependencies across intervening content. We focus our analysis on \suite{Missing Object},\footnote{The modifiers used in the other test suite classes (adjective, ORC, SRC) are not as directly comparable, since they vary in multiple dimensions, not just complexity.} as the modifiers considered in these test suites can be ordered according to their syntactic complexity (no SRC < single SRC < coordinated SRCs < embedded SRCs).

\begin{figure}[t]
\centering
\subfloat[\suite{Garden Path object}\label{fig:gpo}]{
\includegraphics[width=\linewidth]{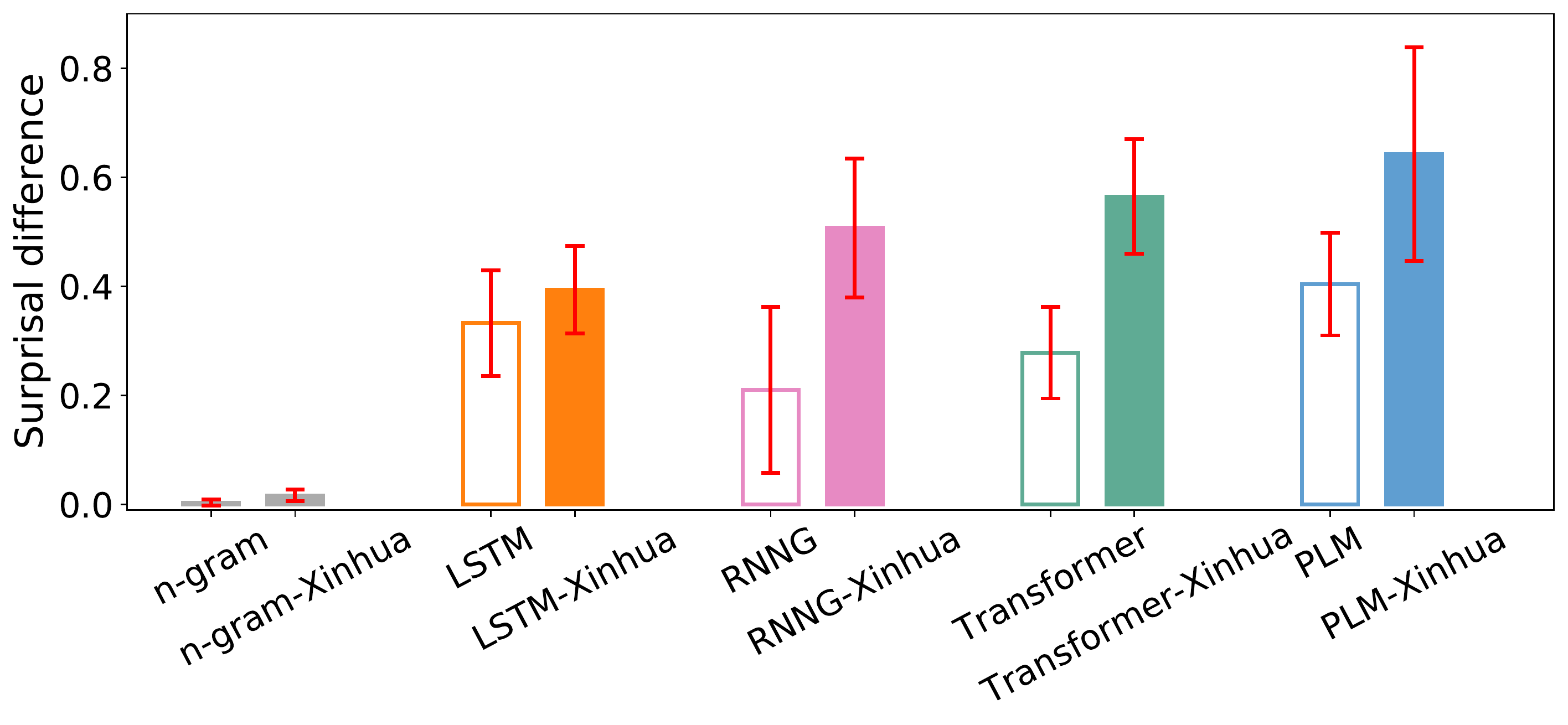}
}
\\
\subfloat[\suite{Garden Path subject}\label{fig:gps}]{
\includegraphics[width=\linewidth]{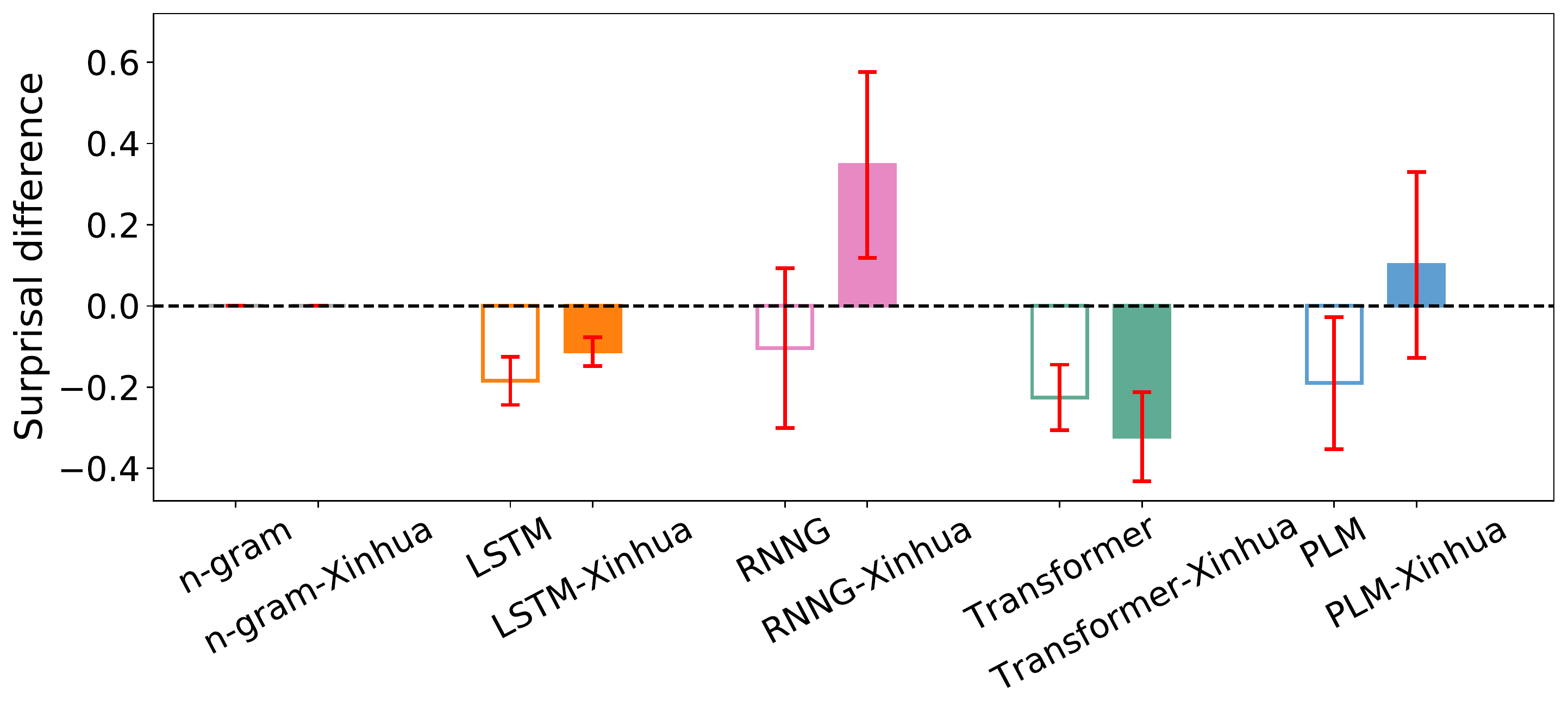}
}
\caption{Mean surprisal difference at target region (matched classifier $-$ mismatched classifier).}
\label{fig:gp}
\end{figure}

\Cref{fig:mobj_line} shows the models' performance on this test suite class as a function of modifier complexity, ranging from least to most difficult along the horizontal axis of each subplot. The vanilla LSTMs and Transformers clearly degrade in performance as the intervening materials between the stimulus and the target region grow in length and complexity. In contrast, it is also visually apparent that the RNNG models do not degrade sharply as modifiers get longer and more complex. We fit linear mixed-effects models to investigate the relationship between modifier type and accuracy for each language model.\footnote{See \Cref{sec:mobj-modifiers} for details.} Our results appear to confirm that both the RNNGs and PLMs do not significantly degrade on the simple SRC and coordinated SRC modifiers (compared to the no-modifier baseline). For the most complex modifier (embedded SRC), all models suffer in accuracy, but the magnitude of this effect is smaller for the RNNGs and PLMs compared to the LSTMs and Transformers. Taken together, our results suggest that while structural supervision does not give the language models a significant advantage compared to their vanilla counterparts in the accuracy scores, it seems to help the model maintain syntactic expectations despite the intervention of syntactically complex content.

\subsection{Garden-Path Effects} \label{sec:gp}

Building upon the \suite{classifier--noun Compatibility} results, we 
investigate whether a mismatched local classifier--noun pair may serve as an early cue for the upcoming RC structure, inducing a garden path effect. \Cref{fig:bymodelntest} shows that the neural models systematically perform better on \suite{Garden Path object} than \suite{Garden Path subject}. We conjecture that neural language models may 
implicitly predict an ORC modifying the subject noun regardless of the type of classifiers.
Therefore, the language models may be more prepared to see an ORC modifying the subject \begin{CJK*}{UTF8}{gbsn}``工厂''\end{CJK*} in (3.b) than in (2.b) with the object \begin{CJK*}{UTF8}{gbsn}``工厂''\end{CJK*}.

To gain a better understanding of model performance on these two test suite classes, we examine the average difference in target-region surprisal values between sentences with and without local classifier--noun mismatch (\Cref{fig:gp}).  
The average difference gives detailed information on how each language model processes the garden path region, which is complementary to the binary success/failure score achieved by a model on a given test item. \Cref{fig:gpo} shows that the neural models have a positive average surprisal difference across test items for \suite{Garden Path object}. Furthermore, the magnitude of this difference increases with the inclusion of the larger Xinhua dataset, suggesting that 
with more data, models become more confident in taking the incongruence between the classifier and the noun as a pre-RC cue.
\footnote{This result also accords with prior findings that classifiers facilitate object-modifying RC processing \cite{wu14,wu2018effects}.} On the other hand, recall that all models perform rather poorly on \suite{Garden Path subject} (\Cref{fig:bymodelntest}). 
\Cref{fig:gps} shows that only the RNNGs trained on the Xinhua corpus output a statistically significant postive average surprisal difference ($p<.001$; one-sample $t$-test). PLM-Xinhua, although not statistically significant, has a positive mean surprisal difference as well. This is due to the fact that the magnitude of the suprisal differences predicted by the RNNG-Xinhua and PLM-Xinhua models is greater when they exhibit the predicted garden-path effects, and smaller when they do not follow the predicted direction. Therefore, structural supervision may help models represent syntactic state in a more human-like way.

\section{Conclusion}
This work evaluates Mandarin Chinese language models on six grammatical relationships, including syntactic dependencies and semantic compatibilities. We use Mandarin as a case study for analyzing how the potential advantages of explicit syntax modeling (as performed by RNNGs and PLMs) generalize from English to a typologically different language.
Although structural supervision does not boost the sequential model’s learning in all relationships tested in this study, we find that it does allow the RNNG and PLM models to learn dependencies robust to increasingly complex modifiers, as seen in the \suite{Missing Object} test suites. Compared to the vanilla sequence-based LSTM and Transformer models, explicit syntactic modeling also seems to help with grammatical generalization in settings with small training data. We also find that Mandarin syntactic dependencies (such as tracking gross syntactic state within a subordinate clause) tend to be easier to learn than semantic dependencies (such as the compatibility between classifiers and nouns). This study is one of the first steps towards understanding the role structural inductive biases may play in learning semantic and syntactic relationships in typologically diverse languages.

\section*{Acknowledgements}
We thank three anonymous reviewers for their feedback. J.H.~acknowledges support from an NSF Graduate Research Fellowship under Grant No.~1745302. This work was supported by the MIT--IBM AI Research Lab and MIT's Quest for Intelligence.

\bibliography{main}
\bibliographystyle{acl_natbib}

\clearpage

\appendix

\section{Individual Test Suites}
\subsection{List of Test Suites}
\label{sec:ind_test_suites}
See Table \ref{tab:summary} for a summary of test suite classes constructed in this study.

\begin{table*}[!t]
\centering
\tiny
\newcommand{\None}{-}
\begin{tabular}{lcccc}
\toprule
\textbf{Test Suite Class} & \textbf{Modifier Type} & \textbf{\# Test Items} & \textbf{Prior Work in English} & \textbf{Prior Work in Chinese}\\
\midrule
\textsc{classifier--noun Compatibility} & None & 30 & \None & \citet{zhan-levy-2018-comparing,xiang2021climp}\\
\textsc{classifier--noun Compatibility} & Adjective & 30 & \None  & \citet{xiang2021climp}\\
\textsc{classifier--noun Compatibility} & Object-Extracted RC & 30 & \None &\citet{xiang2021climp}\\
\textsc{classifier--noun Compatibility} & Subject-Extracted RC & 30 & \None &\citet{xiang2021climp}\\\midrule
\textsc{Garden Path object} & None & 31 & \None &\citet{wu2018effects}\\
\textsc{Garden Path object} & Adjective & 31 & \None & \None\\
\textsc{Garden Path object} & Object-Extracted RC & 31 & \None & \None\\
\textsc{Garden Path object} & Subject-Extracted RC & 31 & \None & \None\\\midrule
\textsc{Garden Path subject} & None & 31 & \citet{futrell-etal-2019-neural, hu-etal-2020-systematic} & \None\\
\textsc{Garden Path subject} & Adjective & 31 & \None & \None\\
\textsc{Garden Path subject} & Object-Extracted RC & 31 & \citet{hu-etal-2020-systematic} & \None\\
\textsc{Garden Path subject} & Subject-Extracted RC & 31 & \citet{hu-etal-2020-systematic} & \None\\\midrule
\textsc{verb--noun Compatibility} & None & 31 & \None & \None\\
\textsc{verb--noun Compatibility} & Adjective & 31 & \None & \None\\
\textsc{verb--noun Compatibility} & Object-Extracted RC & 31 & \None & \None\\
\textsc{verb--noun Compatibility} & Subject-Extracted RC & 31 & \None & \None\\\midrule
\textsc{Missing Object} & None & 30 & \citet{warstadt2020blimp} & \None\\
\textsc{Missing Object} & Subject-Extracted RC & 30 & \None & \None\\
\textsc{Missing Object} & Coordinated Subject-Extracted RC & 30 & \None & \None\\
\textsc{Missing Object} & Embedded Subject-Extracted RC & 30 & \None & \None\\\midrule
\textsc{Subordination} & None & 30 & \citet{futrell-etal-2019-neural,hu-etal-2020-systematic} & \None\\
\textsc{Subordination} & Adjective & 30 & \None & \None\\
\textsc{Subordination} & Object-Extracted RC & 30 & \citet{futrell-etal-2019-neural,hu-etal-2020-systematic} & \None\\
\textsc{Subordination} & Subject-Extracted RC & 30 & \citet{futrell-etal-2019-neural,hu-etal-2020-systematic} & \None\\

\bottomrule
\end{tabular}
\caption{Summary of individual test suites used in our experiments.}
\label{tab:summary}
\end{table*}

\subsection{Complementary Test Suite Examples} \label{sec:modifiers}
In this section, we provide complementary test items for those in \Cref{sec:test-suites}, constructing a full set of examples for each test suite class. Note that except for \suite{Missing Object} where we consider variants of SRCs as the modifier (SRC, coordinated SRC, and embedded SRC), the general modifiers types are adjective, ORC and SRC.
\subsubsection{Classifier--Noun Compatibility} \label{sec:cn-test-suites}
\begin{CJK*}{UTF8}{gbsn}
\modifier{No modifier}
\begin{exe}
\footnotesize
\exi{(7.a)} \gll 孩子 听到 了 一 首 \underline{歌曲\hspace{1mm}。}\\
        child hear \textsc{pst} one \textsc{cl}$_\textsc{song}$ \underline{song .}\\
\glt ``The child heard a song.''
\exi{$*$(7.b)} \gll 孩子 听到 了 一 张  \underline{歌曲\hspace{1mm}。}\\
        child hear \textsc{pst} one \textsc{cl}$_\textsc{album}$ \underline{song .}\\
\glt `The child heard a song.''
\exi{(7.c)} \gll 孩子 听到 了 一 张  \underline{专辑\hspace{1mm}。}\\
        child hear \textsc{pst} one \textsc{cl}$_\textsc{album}$ \underline{album .}\\
\glt ``The child heard an album.''
\exi{$*$(7.d)} 
\gll  孩子 听到 了 一 首 \underline{专辑\hspace{1mm}。}\\
        child hear \textsc{pst} one \textsc{cl}$_\textsc{song}$ \underline{album .}\\
\glt ``The child heard an album.''
\end{exe}
\end{CJK*}
\begin{CJK*}{UTF8}{gbsn}
\modifier{ORC as modifier}
\begin{exe}
\footnotesize
\exi{(8.a)} \gll 孩子 听到 了 一 首 他 熟悉 的 \underline{歌曲\hspace{1mm}。}\\
        child hear \textsc{pst} one \textsc{cl}$_\textsc{song}$ he familiar \textsc{de} \underline{song .}\\
\glt ``The child heard a song that he is familiar with.''
\exi{$*$(8.b)} \gll 孩子 听到 了 一 张 他 熟悉 的 \underline{歌曲\hspace{1mm}。}\\
        child hear \textsc{pst} one \textsc{cl}$_\textsc{album}$ he familiar \textsc{de} \underline{song .}\\
\glt `The child heard a song that he is familiar with.''
\exi{(8.c)} \gll 孩子 听到 了 一 张 他 熟悉 的 \underline{专辑\hspace{1mm}。}\\
        child hear \textsc{pst} one \textsc{cl}$_\textsc{album}$ he familiar \textsc{de} \underline{album .}\\
\glt ``The child heard an album that he is familiar with.''
\exi{$*$(8.d)} 
\gll  孩子 听到 了 一 首 他 熟悉 的 \underline{专辑\hspace{1mm}。}\\
        child hear \textsc{pst} one \textsc{cl}$_\textsc{song}$ he familiar \textsc{de} \underline{album .}\\
\glt ``The child heard an album that he is familiar with.''
\end{exe}
\end{CJK*}
\begin{CJK*}{UTF8}{gbsn}
\modifier{SRC as modifier}
\begin{exe}
\footnotesize
\exi{(9.a)} \gll 孩子 听到 了 一 首 令 他 熟悉 的 \underline{歌曲\hspace{1mm}。}\\
        child hear \textsc{pst} one \textsc{cl}$_\textsc{song}$ make he familiar \textsc{de} \underline{song .}\\
\glt ``The child heard a song that makes him feel familiar.''
\exi{$*$(9.b)} \gll 孩子 听到 了 一 张 令 他 熟悉 的 \underline{歌曲\hspace{1mm}。}\\
        child hear \textsc{pst} one \textsc{cl}$_\textsc{album}$ make he familiar \textsc{de} \underline{song .}\\
\glt `The child heard a song that makes him feel familiar.''
\exi{(9.c)} \gll 孩子 听到 了 一 张 令 他 熟悉 的 \underline{专辑\hspace{1mm}。}\\
        child hear \textsc{pst} one \textsc{cl}$_\textsc{album}$ make he familiar \textsc{de} \underline{album .}\\
\glt ``The child heard an album that makes him feel familiar.''
\exi{$*$(9.d)} 
\gll  孩子 听到 了 一 首 令 他 熟悉 的 \underline{专辑\hspace{1mm}。}\\
        child hear \textsc{pst} one \textsc{cl}$_\textsc{song}$ make he familiar \textsc{de} \underline{album .}\\
\glt ``The child heard an album that makes him feel familiar.''
\end{exe}
\end{CJK*}

\subsubsection{Garden-Path Object}
\begin{CJK*}{UTF8}{gbsn}
\modifier{Adjectival modifier}
\begin{exe}
\footnotesize
\exi{(10.a)} \gll 他 离开 了 那 间 负责 的 朋友 \underline{开} 的 工厂\hspace{1mm}。\\
        he leave \textsc{pst} that \textsc{cl}$_\textsc{building}$ conscientious \textsc{de} friend \underline{start} \textsc{de} factory\hspace{1mm}. \\
\exi{(10.b)} \gll 他 离开 了 那 个 负责 的 朋友 \underline{开} 的 工厂\hspace{1mm}。\\
        he leave \textsc{pst} that \textsc{cl}$_\textsc{general}$ conscientious \textsc{de} friend \underline{start} \textsc{de} factory\hspace{1mm}.\\
\glt ``He left the factory that the conscientious friend started.''
\end{exe}
\end{CJK*}
\begin{CJK*}{UTF8}{gbsn}
\modifier{ORC as modifier}
\begin{exe}
\footnotesize
\exi{(11.a)} \gll 他 离开 了 那 间 我 尊敬 的 朋友 \underline{开} 的 工厂\hspace{1mm}。\\
        he leave \textsc{pst} that \textsc{cl}$_\textsc{building}$ I respect \textsc{de} friend \underline{start} \textsc{de} factory\hspace{1mm}. \\
\exi{(11.b)} \gll 他 离开 了 那 个 我 尊敬 的 朋友 \underline{开} 的 工厂\hspace{1mm}。\\
        he leave \textsc{pst} that \textsc{cl}$_\textsc{general}$ I respect \textsc{de} friend \underline{start} \textsc{de} factory\hspace{1mm}.\\
\glt ``He left the factory that the friend whom I respect started.''
\end{exe}
\end{CJK*}
\begin{CJK*}{UTF8}{gbsn}
\modifier{SRC as modifier}
\begin{exe}
\footnotesize
\exi{(12.a)} \gll 他 离开 了 那 间 帮助 过 我 的 朋友 \underline{开} 的 工厂\hspace{1mm}。\\
        he leave \textsc{pst} that \textsc{cl}$_\textsc{building}$ help \textsc{pst} I \textsc{de} friend \underline{start} \textsc{de} factory\hspace{1mm}. \\
\exi{(12.b)} \gll 他 离开 了 那 个 帮助 过 我 的 朋友 \underline{开} 的 工厂\hspace{1mm}。\\
        he leave \textsc{pst} that \textsc{cl}$_\textsc{general}$ help \textsc{pst} I \textsc{de} friend \underline{start} \textsc{de} factory\hspace{1mm}.\\
\glt ``He left the factory that the friend who helped me before started.''
\end{exe}
\end{CJK*}

\subsubsection{Garden-Path Subject}
\begin{CJK*}{UTF8}{gbsn}
\modifier{Adjectival modifier}
\begin{exe}
\footnotesize
\exi{(13.a)} \gll 那 间 负责 的 朋友 开 \underline{的} 工厂 倒闭 了 。\\
        that \textsc{cl}$_{\textsc{building}}$ conscientious \textsc{de} friend start \underline{\textsc{de}} factory close \textsc{pst} .\\

\exi{(13.b)} \gll 那 个 负责 的 朋友 开 \underline{的} 工厂 倒闭 了 。\\
        that \textsc{cl}$_{\textsc{general}}$ conscientious \textsc{de} friend start \underline{\textsc{de}} factory close \textsc{pst} .\\
\glt ``The factory that the conscientious friend started has closed.''
\end{exe}
\end{CJK*}
\begin{CJK*}{UTF8}{gbsn}
\modifier{ORC as modifier}
\begin{exe}
\footnotesize
\exi{(14.a)} \gll 那 间 我 尊敬 的 朋友 开 \underline{的} 工厂 倒闭 了 。\\
        that \textsc{cl}$_{\textsc{building}}$ I respect \textsc{de} friend start \underline{\textsc{de}} factory close \textsc{pst} .\\

\exi{(14.b)} \gll 那 个 我 尊敬 的 朋友 开 \underline{的} 工厂 倒闭 了 。\\
        that \textsc{cl}$_{\textsc{general}}$ I respect \textsc{de} friend start \underline{\textsc{de}} factory close \textsc{pst} .\\
\glt ``The factory that the friend whom I respect started has closed.''
\end{exe}
\end{CJK*}
\begin{CJK*}{UTF8}{gbsn}
\modifier{SRC as modifier}
\begin{exe}
\footnotesize
\exi{(15.a)} \gll 那 间 帮助 过 我 的 朋友 开 \underline{的} 工厂 倒闭 了 。\\
        that \textsc{cl}$_{\textsc{building}}$ help \textsc{pst} I \textsc{de} friend start \underline{\textsc{de}} factory close \textsc{pst} .\\

\exi{(15.b)} \gll 那 个 帮助 过 我 的 朋友 开 \underline{的} 工厂 倒闭 了 。\\
        that \textsc{cl}$_{\textsc{general}}$ help \textsc{pst} I \textsc{de} friend start \underline{\textsc{de}} factory closed \textsc{pst} .\\
\glt ``The factory that the friend who helped me before started has closed.''
\end{exe}
\end{CJK*}

\subsubsection{Verb--Noun Compatibility}
\begin{CJK*}{UTF8}{gbsn}
\modifier{No modifier}
\begin{exe}
\footnotesize
\exi{(16.a)} \gll 我 修理 了 这 个 \underline{电脑 。} \\
        I fix \textsc{pst} this \textsc{cl} \underline{computer .}\\
        \glt ``I have fixed this computer.''
\exi{$*$(16.b)} \gll 我 阅读 了 这 个 \underline{电脑 。}\\
        I read \textsc{pst} this \textsc{cl} \underline{computer .}\\
\glt ``I have read this computer.''
\end{exe}
\end{CJK*}
\begin{CJK*}{UTF8}{gbsn}
\modifier{ORC as modifier}
\begin{exe}
\footnotesize
\exi{(17.a)} \gll 我 修理 了 这 个 父亲 使用 的 \underline{电脑 。} \\
        I fix \textsc{pst} this \textsc{cl} father use \textsc{de} \underline{computer .}\\
        \glt ``I have fixed this computer that the father uses.''

\exi{$*$(17.b)} \gll 我 阅读 了 这 个 父亲 使用 的 \underline{电脑 。}\\
        I read \textsc{pst} this \textsc{cl} father use \textsc{de} \underline{computer .}\\
\glt ``I have read this computer that the father uses.''
\end{exe}
\end{CJK*}
\begin{CJK*}{UTF8}{gbsn}
\modifier{SRC as modifier}
\begin{exe}
\footnotesize
\exi{(18.a)} \gll 我 修理 了 这 个 计算 公式 的 \underline{电脑 。} \\
        I fix \textsc{pst} this \textsc{cl} calculate formula \textsc{de} \underline{computer .}\\
        \glt ``I have fixed this computer that calculates formulas.''
\exi{$*$(18.b)} \gll 我 阅读 了 这 个 计算 公式 的 \underline{电脑 。}\\
        I read \textsc{pst} this \textsc{cl} calculate formula \textsc{de} \underline{computer .}\\
\glt ``I have read this computer that calculates formulas.''
\end{exe}
\end{CJK*}

\subsubsection{Missing Object} \label{sec:mobj-test-suites}
\begin{CJK*}{UTF8}{gbsn}
\modifier{SRC as modifier}
\begin{exe}
\footnotesize
\exi{(19.a)} 
\gll 记者 采访 了 研发 产品 的 科学家 \underline{。} \\
        journalist interview \textsc{pst} develop product \textsc{de} scientist \underline{.}\\
\glt ``The journalist interviewed the scientist who developed the product.''
\exi{$*$(19.b)} 
\gll 记者 采访 了 研发 产品 \underline{。} \\
        journalist interview \textsc{pst} develop product \underline{.}\\
\glt ``The journalist interviewed who developed the product.''
\end{exe}
\end{CJK*}
\begin{CJK*}{UTF8}{gbsn}
\modifier{Coordinated SRCs as modifier}
\begin{exe}
\footnotesize
\exi{(20.a)} \gll 记者 采访 了 研发 产品 并且 获 了 奖 的 科学家 \underline{。}\\
        journalist interview \textsc{pst} develop product and win \textsc{pst} prize \textsc{de} scientist \underline{.}\\
\glt ``The journalist interviewed the scientist who developed the product and won a prize.''
\exi{$*$(20.b)} \gll 记者 采访 了 研发 产品 并且 获 了 奖 \underline{。}\\
        journalist interview \textsc{pst} develop product and win \textsc{pst} prize \underline{.}\\
\glt ``The journalist interviewed who developed the product and won a prize.''
\end{exe}
\end{CJK*}
\begin{CJK*}{UTF8}{gbsn}
\modifier{Embedded RCs as modifier}
\begin{exe}
\footnotesize
\exi{(21.a)} \gll 记者 采访 了 研发 帮助 老人 的 产品 的 科学家 \underline{。}\\
        journalist interview \textsc{pst} develop help elderly \textsc{de} product \textsc{de} scientist \underline{.}\\
\glt ``The journalist interviewed the scientist who developed the product that helps the elderly.''
\exi{$*$(21.b)} \gll 记者 采访 了 研发 帮助 老人 的 产品 \underline{。}\\
        journalist interview \textsc{pst} develop help elderly \textsc{de} product \underline{.}\\
\glt ``The journalist interviewed who developed the product that helps the elderly.''
\end{exe}
\end{CJK*}

\subsubsection{Subordination}
\begin{CJK*}{UTF8}{gbsn}
\modifier{Adjectival modifier}
\begin{exe}
\footnotesize
\exi{(22.a)} \gll 如果 内向 的 他 不 尝试 ， 他 将 失去 机会 \underline{。}\\
        if introverted \textsc{de} he \textsc{neg} try , he will lose opportunity \underline{.}\\
\glt ``If he who is introverted doesn't try, he will lose the opportunity.''
\exi{$*$(22.b)} \gll 如果 内向 的 他 不 尝试 \underline{。}\\
        if introverted \textsc{de} he \textsc{neg} try \underline{.} \\
\glt ``If he who is introverted doesn't try.''
\end{exe}
\end{CJK*}
\begin{CJK*}{UTF8}{gbsn}
\modifier{ORC as modifier}
\begin{exe}
\footnotesize
\exi{(23.a)} \gll 如果 父亲 期待 的 他 不 尝试 ， 他 将 失去 机会 \underline{。}\\
        if father expect \textsc{de} he \textsc{neg} try , he will lose opportunity \underline{.}\\
\glt ``If he who the father has expectations of doesn't try, he will lose the opportunity.''
\exi{$*$(23.b)} \gll 如果 父亲 期待 的 他 不 尝试 \underline{。}\\
        if father expect \textsc{de} he \textsc{neg} try \underline{.} \\
\glt ``If he who the father has expectations of doesn't try.''
\end{exe}
\end{CJK*}
\begin{CJK*}{UTF8}{gbsn}
\modifier{SRC as modifier}
\begin{exe}
\footnotesize
\exi{(24.a)} \gll 如果 没有 工作 的 他 不 尝试 ， 他 将 失去 机会 \underline{。}\\
        if \textsc{neg} job \textsc{de} he \textsc{neg} try , he will lose opportunity \underline{.}\\
\glt ``If he who has no jobs doesn't try, he will lose the opportunity.''
\exi{$*$(24.b)} \gll 如果 没有 工作 的 他 不 尝试 \underline{。}\\
        if \textsc{neg} job \textsc{de} he \textsc{neg} try \underline{.} \\
\glt ``If he who has no jobs doesn't try.''
\end{exe}
\end{CJK*}

\section{Model Information}
\label{sec:model-info-ppl}

\begin{table}[ht]
\centering
\small

\subfloat[\label{tab:model}]{
\begin{tabular}{lccc}
\toprule
\textbf{Model} & \textbf{\# layers} & \textbf{\# hidden units} & \textbf{Emb size}\\
\midrule
LSTM & 2 & 256 & 256 \\
RNNG & 2 & 256 & 256 \\
Transformer & 12 & 768 & 768 \\ 
PLM & 12 & 768 & 768 \\ \bottomrule
\end{tabular}}

\subfloat[\label{tab:ppl}]{
\begin{tabular}{lcc} \toprule
\textbf{Model} & \textbf{CTB} & \textbf{Xinhua}\\
\midrule
$n$-gram & 330 & 332 \\
LSTM & 161 & 190 \\
RNNG & 227 & 194 \\
Transformer & 234 & 170 \\
PLM & 297 & 244 \\\bottomrule
\end{tabular}
}
\caption{(a) Model architecture size. (b) Perplexity results on CTB test data.} \label{tab:models_corpora}
\end{table}

We find that the perplexity score reported in \Cref{tab:ppl} is comparatively high for RNNG compared to that reported in \citet{dyer2016recurrent}. This may be because the CTB data we use includes some informal and spoken language, such as weblogs and broadcast conversations.

\section{Corpus Statistics}
\label{sec:corpus-stats}

\begin{table}[ht]
    \centering
    \small
    \begin{tabular}{lcc}
    \toprule
    \textbf{Corpus} & \textbf{\# Tokens} & \textbf{Vocab Size}\\
    \midrule
    CTB & 974K & 27K \\
    Xinhua & 7M & 88K \\\bottomrule
    \end{tabular}
    \caption{Statistics of training corpora.}
    \label{tab:corpus-stats}
\end{table}

\begin{figure}[!ht]
    \centering
    \includegraphics[width=\linewidth]{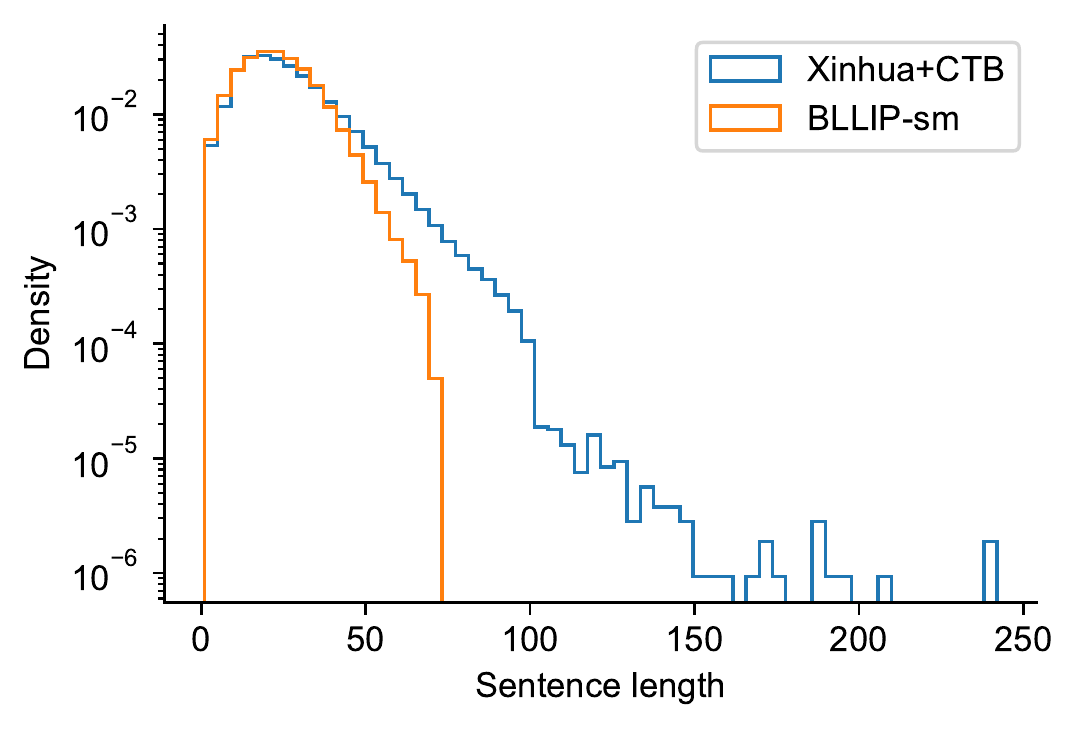}
    \caption{Comparing the distribution of sentence length in Mandarin and English corpora.}
    \label{fig:sent-len-stats}
\end{figure}

Figure \ref{fig:sent-len-stats} shows differences in the distribution of sentence lengths between the Mandarin corpus and a comparable English corpus of similar register. The maximum sentence length is 242 words for Xinhua+CTB, and only 70 words for the English BLLIP-sm corpus used to train \citeauthor{hu-etal-2020-systematic}'s (\citeyear{hu-etal-2020-systematic}) models. The distribution of sentence lengths also has a heavier tail for the Mandarin corpus than the English corpus.

\section{Additional Analysis} \label{sec:add-analysis}
\subsection{Comparing Syntactic and Semantic Test Suites}
\label{sec:compare-syntactic-semantic}

In this section, we focus on comparing how well models learn about syntactic dependency and semantic compatibility.

\begin{figure}[!ht]
\includegraphics[width=\linewidth]{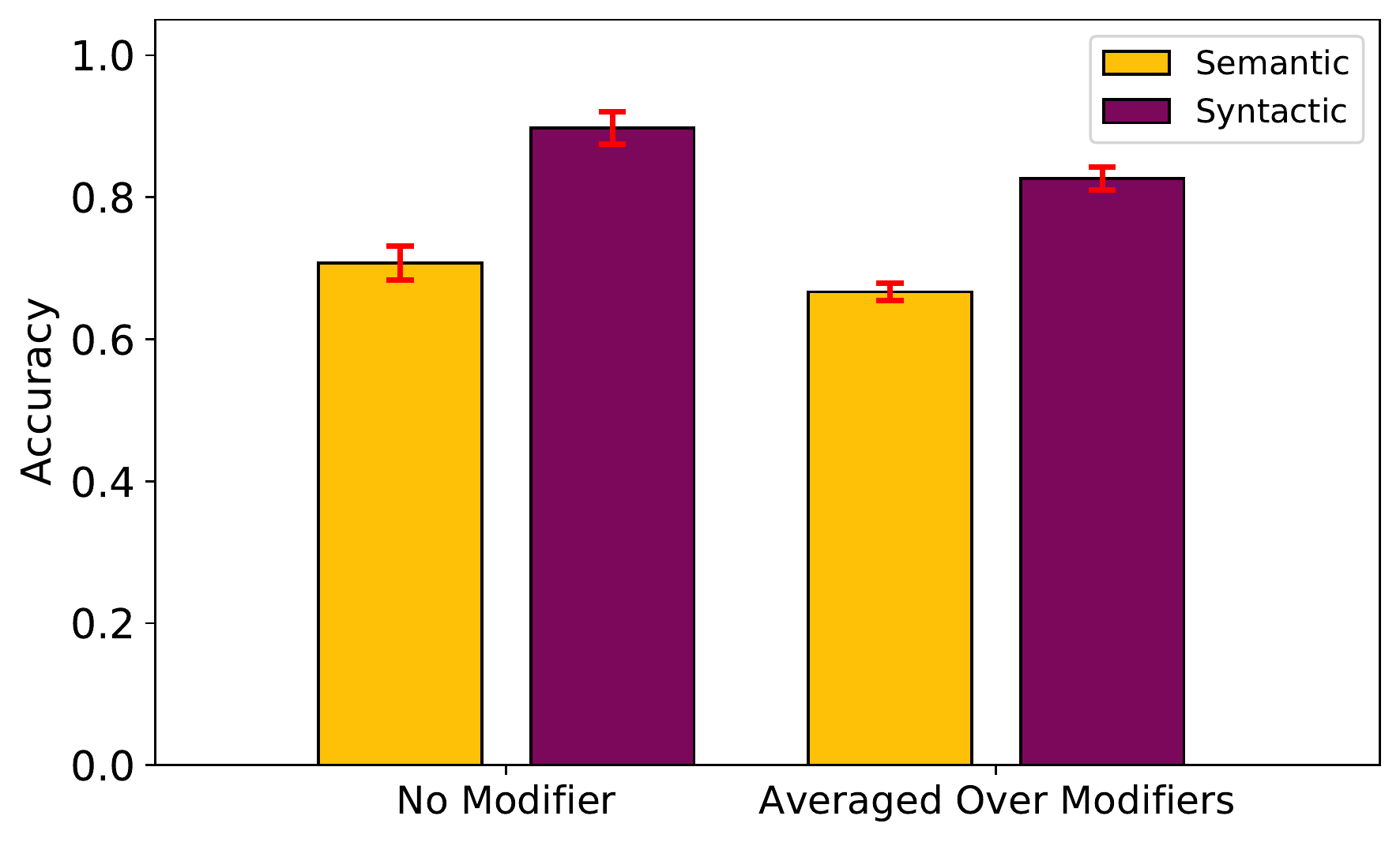}
\caption{Accuracy on semantic and syntactic test suite classes, with and without intervening content. $n$-gram results are excluded.} 
    \label{fig:synsem}
\end{figure}

Recall that we group \suite{classifier--noun Compatibility} and \suite{verb--noun Compatibility} as semantic relationships, and 
\suite{Missing Object} and \suite{Subordination} as syntactic dependency. We compute the accuracy scores of these two categories, as shown in Figure \ref{fig:synsem}. Here we exclude the $n$-gram models since their performances deviate from the other language models to a great extent. Adding modifiers between the stimulus and the target region seems to shrink the gap a bit. And this again might be due to the fact that for \suite{Missing Object}, we intentionally make the intervening contents increasingly hard to learn, dragging down the average accuracy score for syntactic dependencies.

Similar to testing the effect of data size, we determine the statistical significance here with a linear mixed-effects model on the accuracy score. We predict accuracy with a binary indicator of whether or not the test is in the syntactic group that we define, including model type and test item as random factors. We find that syntactic relationships seem to be easier to learn than semantic ones regardless of the intervening content ($p<.001$).

\subsection{Intervening Content in \suite{Missing Object}}
\label{sec:mobj-modifiers}
In this section, we discuss our statistical analysis on language models' robustness against intervening contents.

We fit separate linear mixed-effects models for each language model with the accuracy score as the dependent variable, and modifier type as the predictor. We include model seed and data size as random factors, both with random intercepts and random slopes. \Cref{tab:mobj_coef} summarizes our results. Recall that in \suite{Missing Object}, we consider four types of modifiers: None, single SRC, coordinated SRCs, and embedded SRCs. Each cell in \Cref{tab:mobj_coef} represents coefficients of a particular modifier with respect to the None modifier baseline for a particular language model class. For single SRC and coordinated SRC modifiers, neither of the RNNG or PLM models show significant degradation in the accuracy score. All models suffer from the embedded SRC modifier, with negative coefficients that are all statistically significant. However, RNNGs and PLMs seem to be affected the least, with larger coefficients than vanilla LSTMs and Transformers. This suggests that structural supervision helps language model to learn syntactic dependencies that are more robust against the intervening content.

\begin{table}[ht]
\centering
\scriptsize

\begin{tabular}{llll}
\toprule
\textbf{Model} & \textbf{Single SRC} & \textbf{Coordinated SRC} & \textbf{Embedded SRC}\\
\midrule
LSTM & -0.06* & -0.14*** & -0.3*** \\
RNNG & -0.016 & -0.025 & -0.175*** \\
Transformer & -0.1** & -0.13* & -0.35*** \\ 
PLM & 0.05* & -0.01 & -0.2*** \\ \bottomrule
\end{tabular}

\caption{Coefficients of the modifiers in \suite{Missing Object} with the condition of no modifier as the baseline. *: $p\leq .05$, **: $p\leq .01$, ***: $p\leq .001$.} \label{tab:mobj_coef}
\end{table}

\section{Accuracy by Model and Test Suite Class} \label{sec:table_acc}
See Table \ref{tab:model_results} for a summary of accuracy scores achieved by each model across test suite classes.

\begin{table*}[t]
\centering
\scriptsize

\begin{tabular}{lccccccc}
\toprule
\textbf{Model} & \suite{classifier--noun} & \begin{tabular}{@{}c@{}}\suite{Garden Path} \\ \suite{object}\end{tabular} & \begin{tabular}{@{}c@{}}\suite{Garden Path} \\ \suite{subject}\end{tabular} & \suite{verb--noun} & \suite{Missing Object} & \suite{Subordination}\\
\midrule
$n$-gram & 0.552 & 0.508 & \textcolor{blue}{0.484} & 0.484 & 0.575 & 0.675 \\
LSTM  & 0.598 & 0.659 & 0.320 & 0.624 & \textcolor{blue}{0.847} & 0.789 \\
RNNG  & 0.609 & 0.690 & 0.359 & \textcolor{blue}{0.714} & 0.838 & \textcolor{blue}{0.854}\\
Transformer & 0.603 & 0.672 & 0.376 & 0.581 & 0.761 & 0.817 \\ 
PLM  & \textcolor{blue}{0.615} & \textcolor{blue}{\textbf{0.750}} & 0.419 & 0.589 & 0.775 & 0.836 \\
\midrule
$n$-gram-Xinhua  & 0.594 & 0.516 & \textcolor{black}{\textbf{0.516}} & 0.500 & 0.700 & 0.500 \\
LSTM-Xinhua & 0.650 & 0.691 & 0.355 & \textcolor{black}{\textbf{0.782}} & 0.819 & 0.850 \\
RNNG-Xinhua & 0.636 & \textcolor{black}{\textbf{0.750}} & 0.367 & 0.714 & \textcolor{black}{\textbf{0.854}} & 0.913 \\
Transformer-Xinhua & 0.708 & 0.720 & 0.363 & 0.755 & 0.792 & \textcolor{black}{\textbf{0.931}} \\ 
PLM-Xinhua & \textcolor{black}{\textbf{0.746}} & 0.656 & 0.395 & 0.745 & 0.692 & 0.908 \\
\bottomrule
\end{tabular}\hfill%

\caption{Accuracy score by model and test suite class. Blue color denotes the best score within the CTB dataset.}
\label{tab:model_results}
\end{table*}

\section{Results on Individual Test Suites}
\label{sec:results_ind_test_suites}
\begin{figure*}[t]
\centering
\subfloat[Classifier--Noun Compatibility]{
\includegraphics[width=0.49\linewidth]{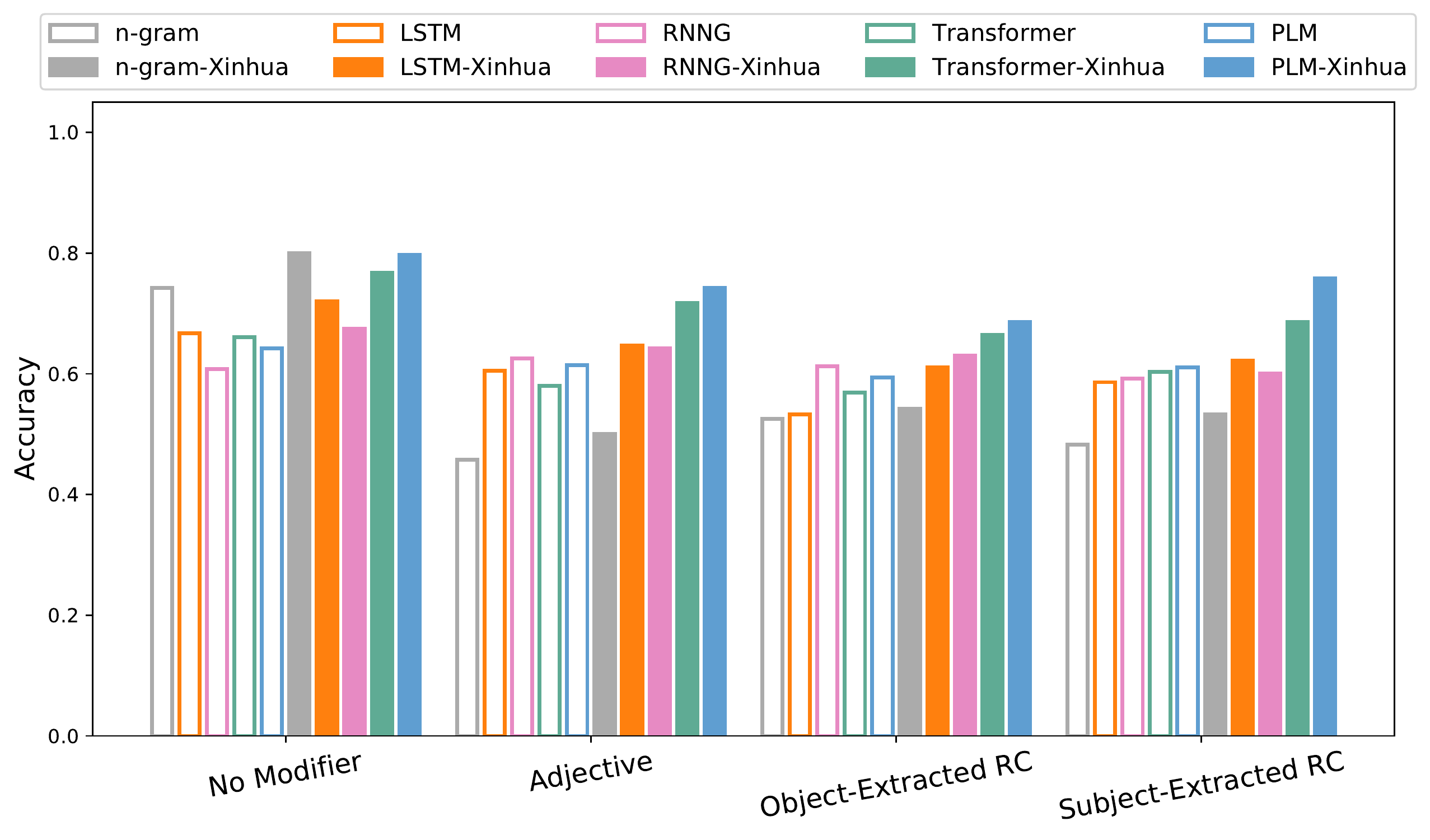}
}
\subfloat[Verb--Noun Compatibility]{
\includegraphics[width=0.49\linewidth]{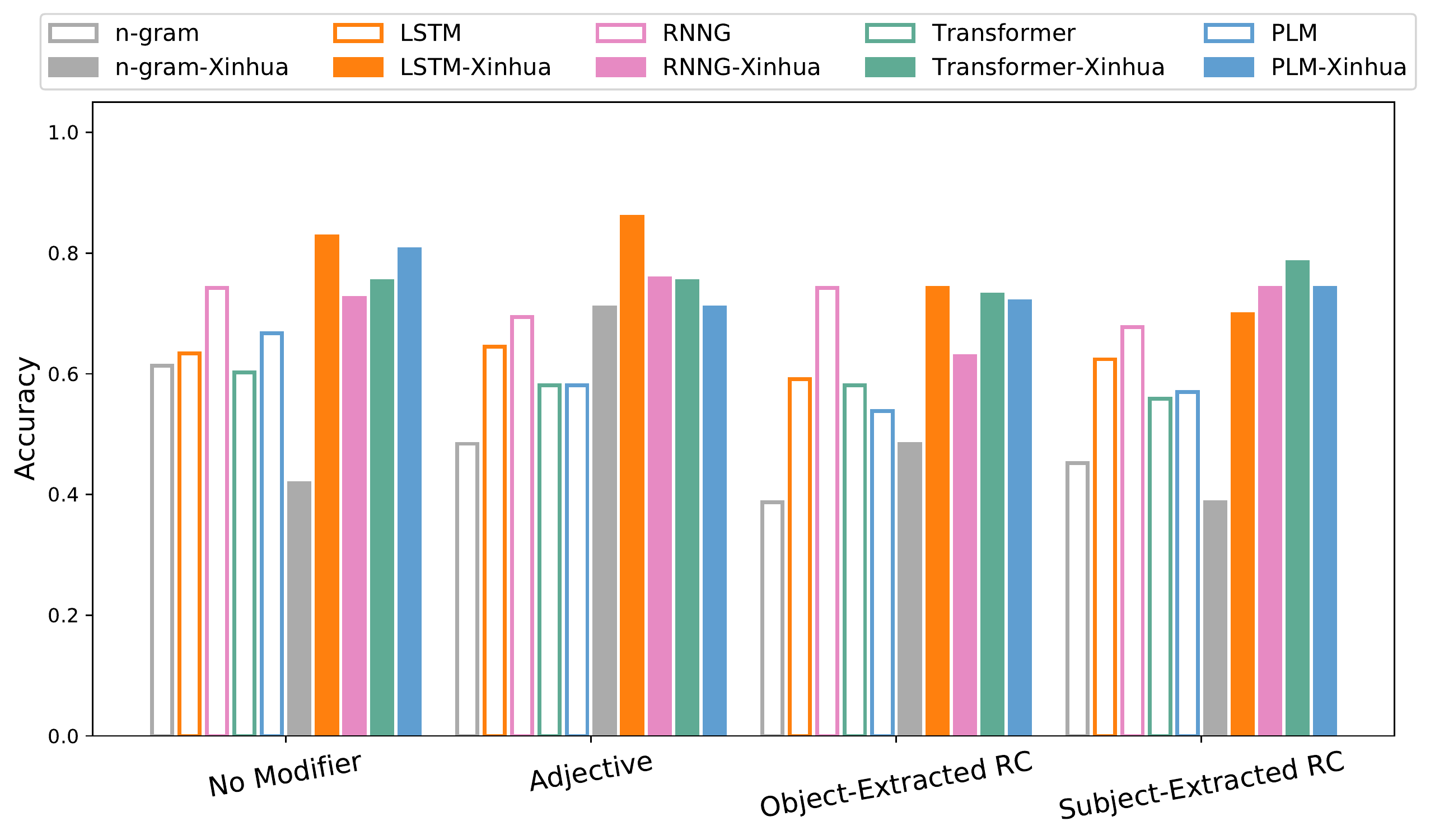}
}

\subfloat[Garden Path Object]{
\includegraphics[width=0.49\linewidth]{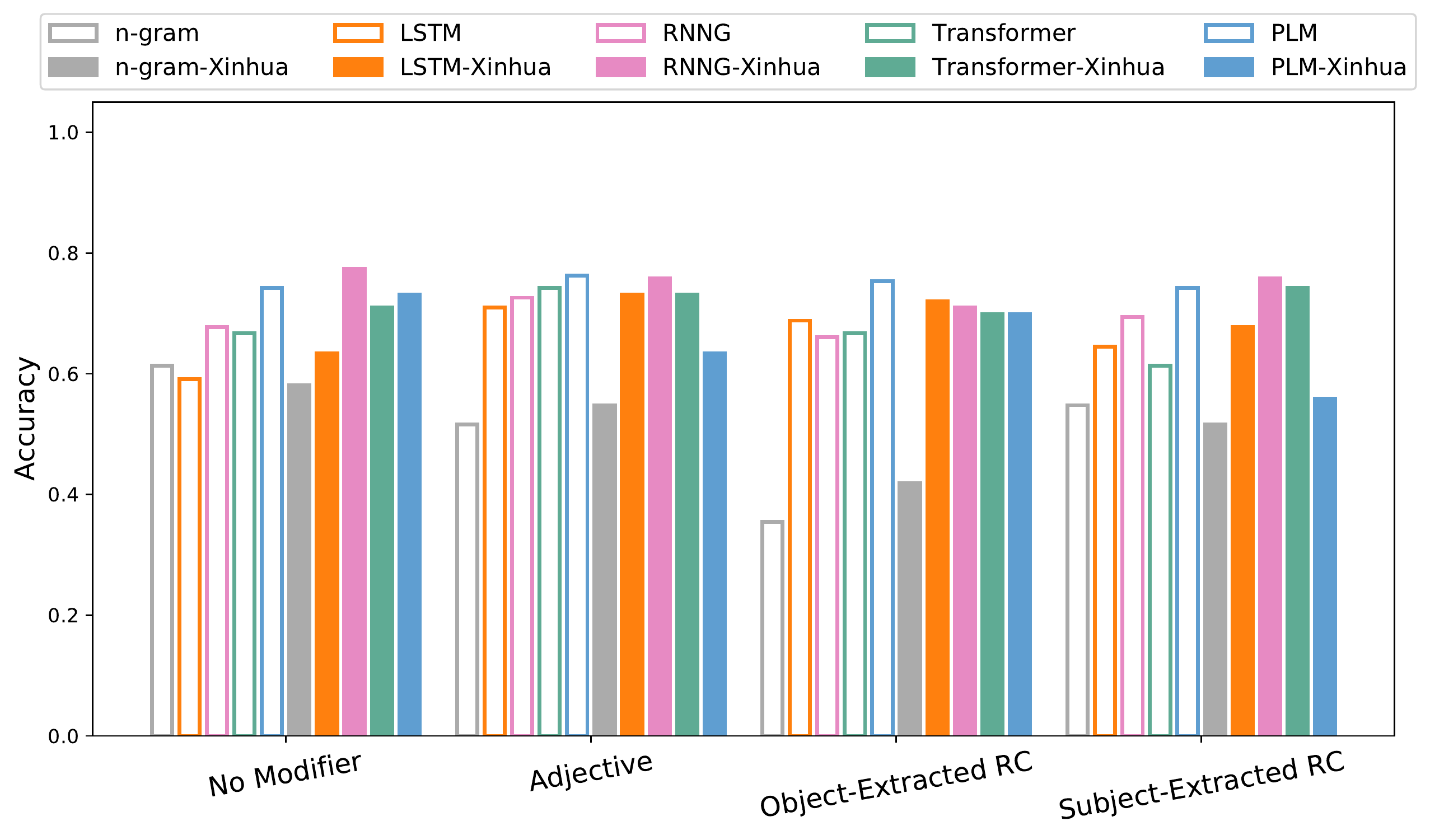}
}
\subfloat[Garden Path Subject]{
\includegraphics[width=0.49\linewidth]{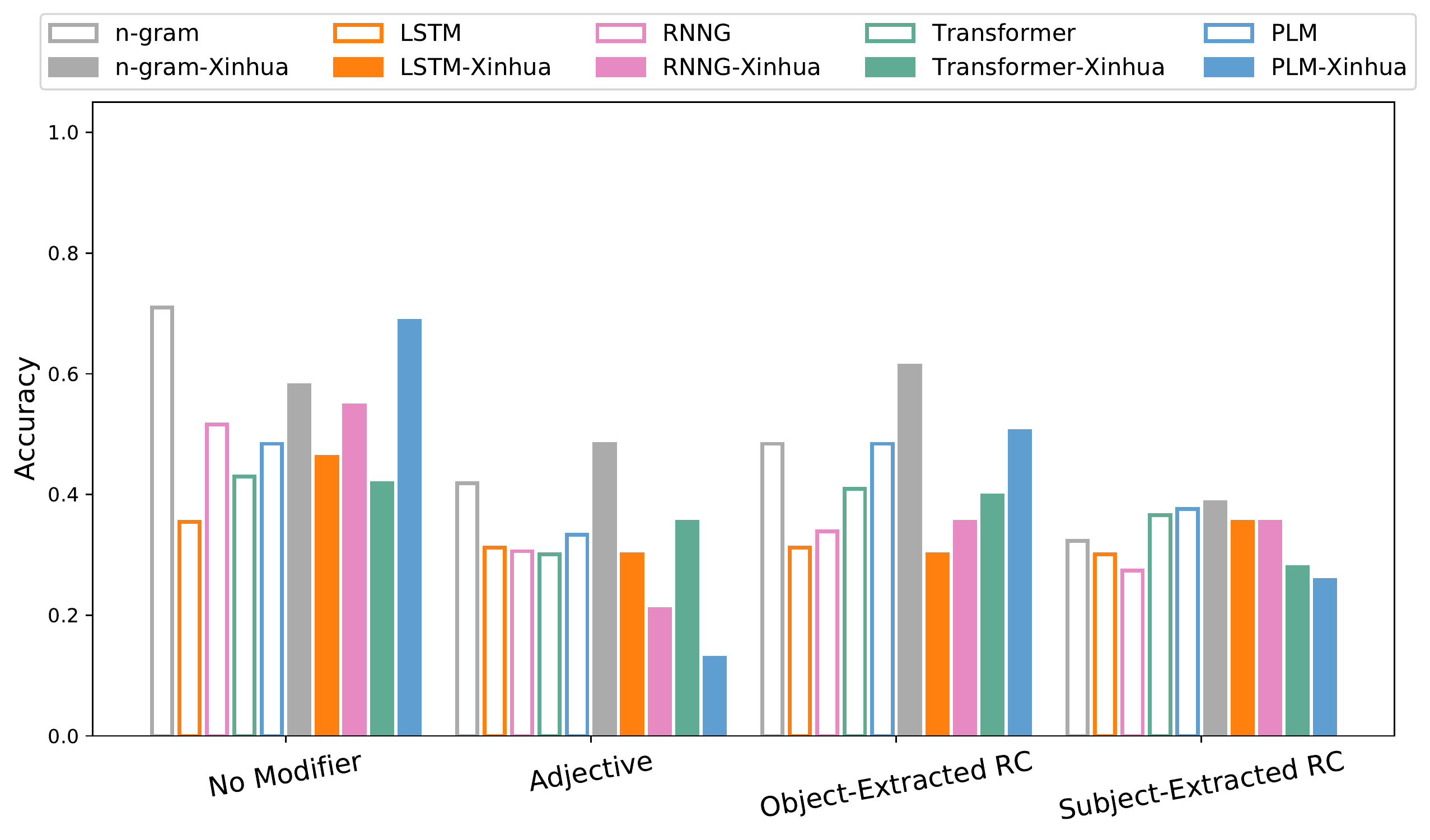}
}

\subfloat[Missing Object]{
\includegraphics[width=0.49\linewidth]{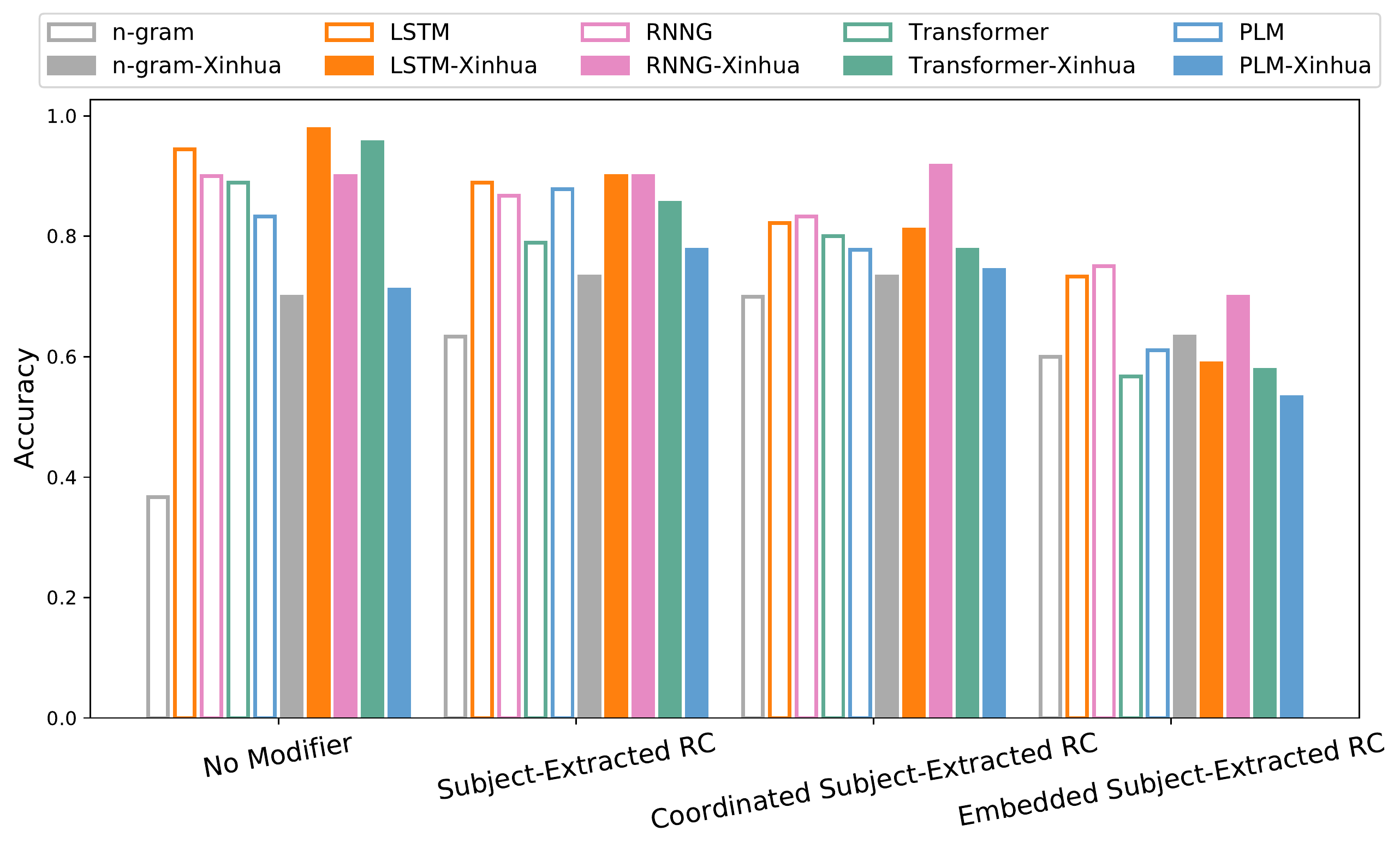}
}
\subfloat[Subordination]{
\includegraphics[width=0.49\linewidth]{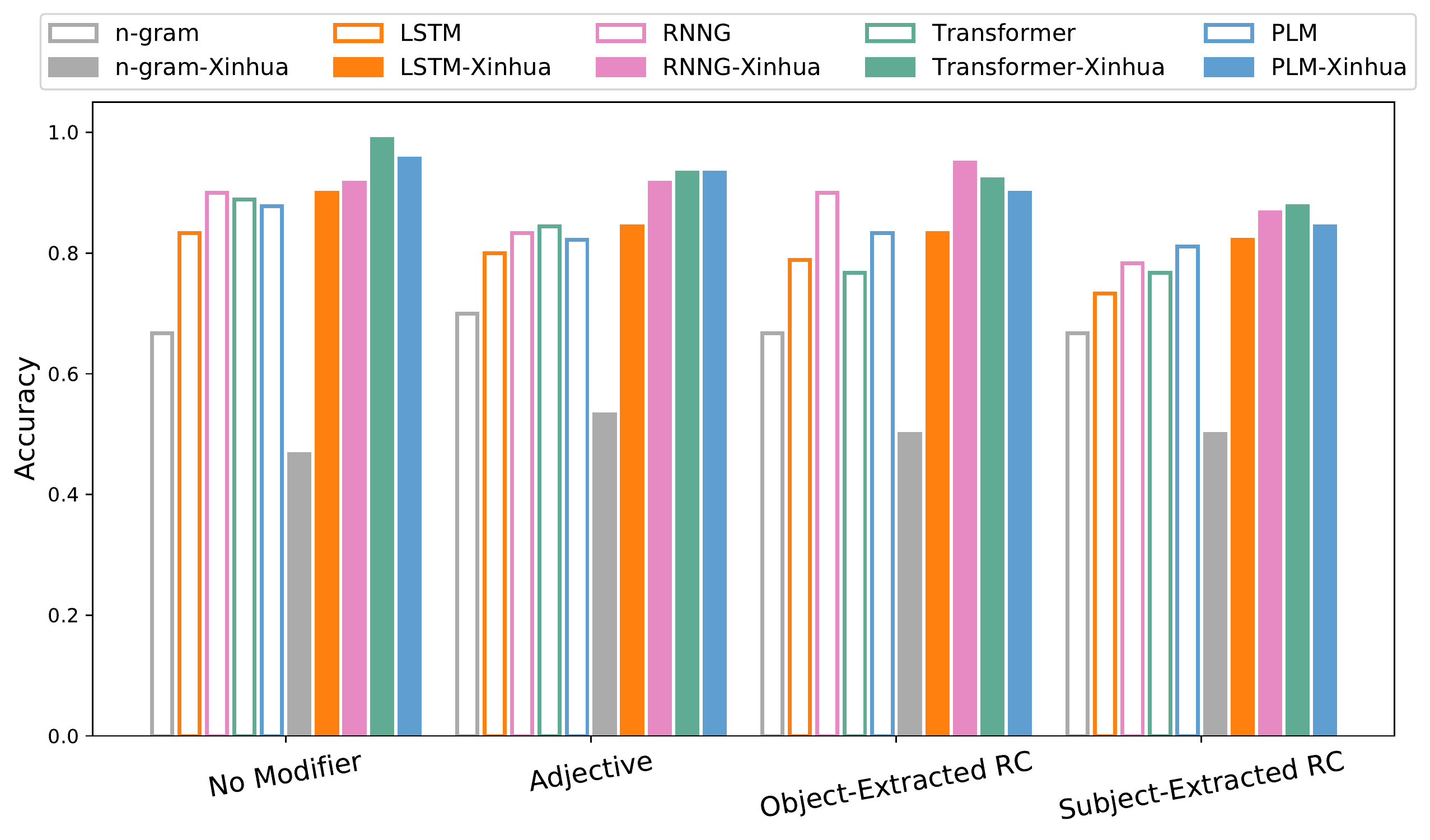}
}
\caption{Accuracy on individual test suites used in our experiments.}
\label{fig:append}
\end{figure*}

\Cref{fig:append} shows language models' accuracy scores on the six test suite classes: \suite{classifier--noun Compatibility}, \suite{Garden Path object}, \suite{Garden Path subject}, \suite{verb--noun Compatibility}, \suite{Missing Object}, and \suite{Subordination}. On the x-axis, we have four types of modifiers tested on that specific test suite class.

\end{document}